%% file: AIPQ_arxiv.tex
\documentclass{bmvc2k}
\usepackage{graphicx}
\usepackage{tikz}
\usepackage{comment}
\usepackage{amsmath,amssymb} 
\usepackage{color}

\usepackage[accsupp]{axessibility}  

\usepackage{bbm}
\usepackage{dirtytalk}
\usepackage{pifont}
\usepackage{overpic}
\usepackage{multirow}
\usepackage{tabulary}
\usepackage{comment}
\usepackage{cuted}
\usepackage{enumitem}

\newlength\savewidth\newcommand\shline{\noalign{\global\savewidth\arrayrulewidth
  \global\arrayrulewidth 1pt}\hline\noalign{\global\arrayrulewidth\savewidth}}
\newcommand{\tablestyle}[2]{\setlength{\tabcolsep}{#1}\renewcommand{\arraystretch}{#2}\centering\footnotesize} 

\DeclareMathOperator{\rank}{rank}
\DeclareMathOperator{\sign}{sgn}
\DeclareMathOperator{\cov}{cov}

\usepackage[labeled,resetlabels]{multibib} 
\newcites{S}{References} 

\usepackage{xspace}
\usepackage{cleveref}
\usepackage{tabularx}
\usepackage{xcolor}
\usepackage{forloop}
\usepackage{adjustbox}

\makeatletter
\DeclareRobustCommand\onedot{\futurelet\@let@token\@onedot} 
\def\@onedot{\ifx\@let@token.\else.\null\fi\xspace}

\def\eg{\emph{e.g}\onedot} 
\def\ie{\emph{i.e}\onedot} 
\def\cf{\emph{c.f}\onedot} 
 \def\vs{\emph{vs}\onedot}
 
\def\etal{\emph{et al}\onedot}
\makeatother

\makeatletter
\renewcommand\paragraph{\@startsection{paragraph}{4}{\z@}%
  {.2em \@plus1ex \@minus.5ex}
  {-.5em}%
  {\normalfont\normalsize\bfseries\textcolor{bmv@captioncolor}}}
\makeatother

\title{Content-Diverse Comparisons improve IQA}

\addauthor{William Thong \textsuperscript{$\ast$} }{william.thong@sony.com}{1,2}
\addauthor{Jose Costa Pereira}{jose.c.pereira@huawei.com}{3}
\addauthor{Sarah Parisot}{sarah.parisot@huawei.com}{3}
\addauthor{Ales Leonardis}{ales.leonardis@huawei.com}{3}
\addauthor{Steven McDonagh}{steven.mcdonagh@huawei.com}{3}

\addinstitution{University of Amsterdam\\The Netherlands}
\addinstitution{Sony AI\\Switzerland}
\addinstitution{Huawei Noah's Ark Lab\\United Kingdom}

\runninghead{Thong~\etal}{Content-Diverse Comparisons improve IQA}

\begin{document}

\maketitle

\begin{abstract}
Image quality assessment (IQA) forms a natural and often straightforward undertaking for humans, yet effective automation of the task remains highly challenging.
Recent metrics from the deep learning community commonly compare image pairs during training to improve upon traditional metrics such as PSNR or SSIM.
However, current comparisons ignore the fact that image content affects quality assessment as comparisons only occur between images of similar content.
This restricts the diversity and number of image pairs that the model is exposed to during training.
In this paper, we strive to enrich these comparisons with content diversity.
Firstly, we relax comparison constraints, and compare pairs of images with differing content. This increases the variety of available comparisons.
Secondly, we introduce listwise comparisons to provide a holistic view to the model. By including differentiable regularizers, derived from correlation coefficients, models can better adjust predicted scores relative to one another.
Evaluation on multiple benchmarks, covering a wide range of distortions and image content, shows the effectiveness of our learning scheme for training image quality assessment models.
\end{abstract}

\section{Introduction}
\label{sec:intro}

\begin{figure}[t]
\begin{minipage}{0.35\textwidth}
\centering
\hfill
\begin{overpic}[width=0.9\linewidth]{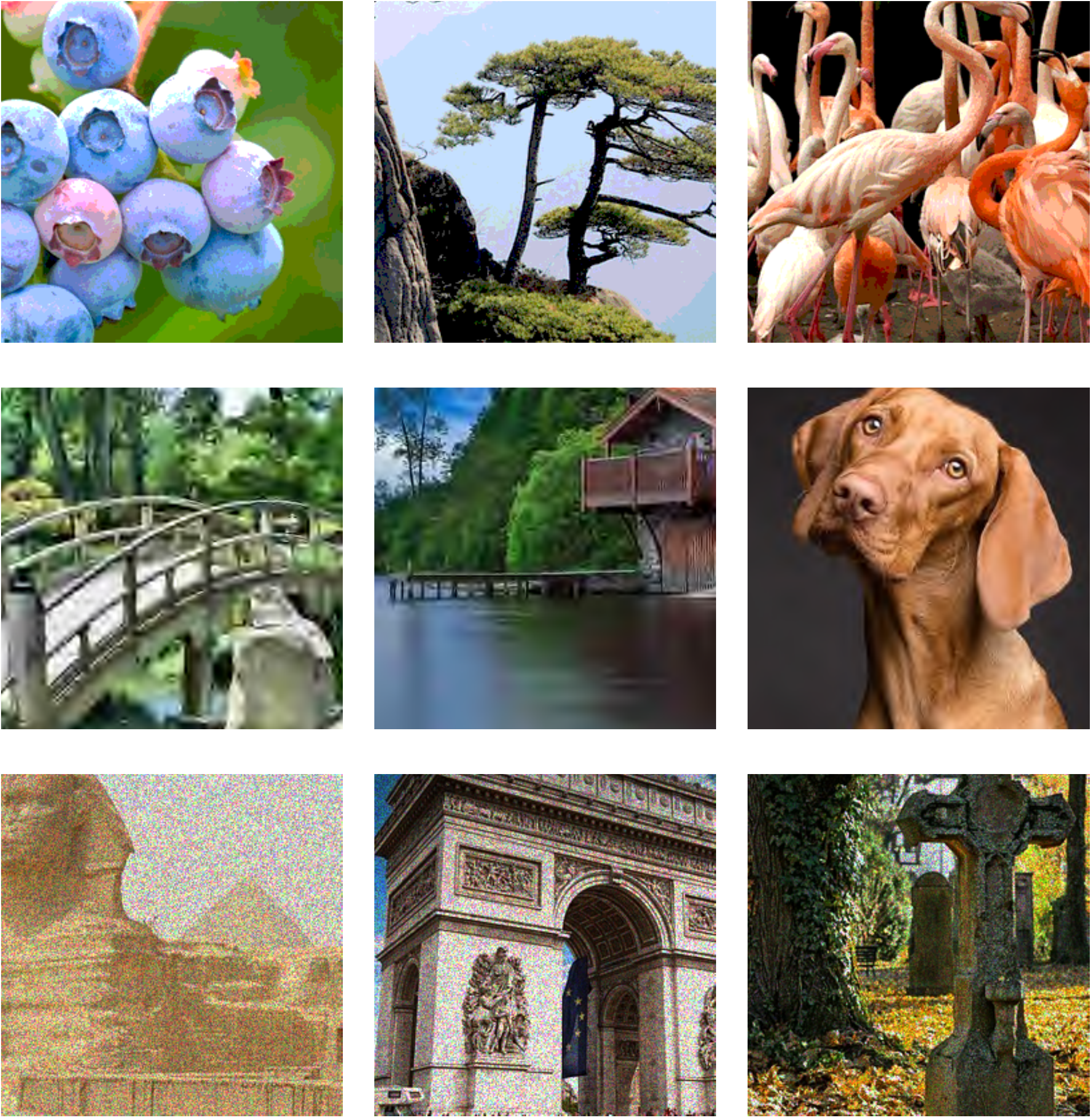}

\put(-5,73){\tiny\rotatebox{90}{Quantization}}
\put(-10,30){\tiny\rotatebox{90}{\parbox{50px}{\centering JPEG2000\\compression}}}
\put(-10,-5){\tiny\rotatebox{90}{\parbox{50px}{\centering Multiplicative\\noise}}}

\put(4,-4){\tiny \textit{Low}-quality}
\put(35,-4){\tiny \textit{Medium}-quality}
\put(71,-4){\tiny \textit{High}-quality}

\end{overpic}
\end{minipage}
\hfill
\begin{minipage}{0.6\textwidth}
\caption{
\textbf{Image Quality Assessment} by human judgement (columns) is affected by the image content despite a similar distortion severity and type (rows).
Enabling models to incorporate this observation leads to richer training signals.
\label{fig:teaser}}
\end{minipage}
\vspace{-1em}
\end{figure}
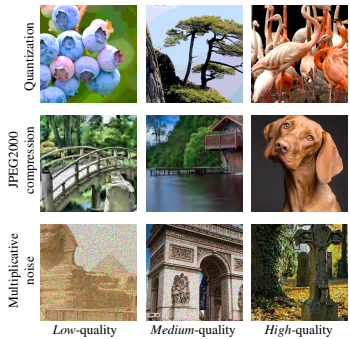

This paper addresses image quality assessment, where the task is to assign quantitative scores to rank images by their perceptual quality.
An ability to perform this task reliably and accurately is of critical importance to assess real-world applications such as image compression~\cite{ponomarenko2015tid,larson2010csiq,sheikh2003live}, restoration~\cite{liu2013deblurring,min2019dehazing} or synthesis~\cite{ma2017learning,tian2018rendering}.
{For example, visual artifacts can appear when applying an image compression algorithm, creating distortions that affect quality in comparison with an original pristine version. Towards robustly measuring image quality, the gold standard involves providing human observers the task of independently grading images on the axis of perceptual quality. A mean opinion score, per image, then provides relative rankings of images according to their respective average ratings and therefore perceived quality~\cite{sheikh2006statistical,ponomarenko2013color}.}

Recent methods typically approach the ranking problem by learning to compare a pair of distorted images, originating from similar content~\cite{liu2017rankiqa, prashnani2018pieapp, zhang2018lpips, ding2021dists}.
However, this ignores the fact that image content often plays an important role when determining human perceived image quality~\cite{siahaan2016does, siahaan2016augmenting, siahaan2018semantic, watson1997digital}.
As illustrated in Fig.~\ref{fig:teaser}, for a similar distortion severity and type, the perceived image quality varies with the image content. In other words, a given distortion affects perceptual quality differently, depending on the image content.
For example, there can exist two images with differing content and differing distortions but with a similar mean opinion score; or alternatively two images with a similar distortion and differing content but with differing mean opinion scores.
As such, we argue that models could gain richer training signals by going beyond comparing images with similar content, and start comparing images with differing content.

Relying on \emph{pairwise} comparisons with similar image content simplifies the image quality ranking task~\cite{liu2017rankiqa, prashnani2018pieapp}.
Indeed, models need only to compare two different distortions.
Nevertheless, this formulation on image content can become a constraint as it doesn't reflect the observation in Fig.~\ref{fig:teaser} and restricts the number of valid image pairs available to sample during training.
Thus, we propose to generalize pairwise comparisons by considering all possible pairs, regardless of their image content, during model training.
This relaxation provides more content diversity in the pairwise comparisons, which in return better leverages the available ground truth image quality annotations.

As comparisons are key to model training, we further explore \emph{listwise} comparisons.
While commonly employed in information retrieval settings~\cite{li2014learning, liu2011learning}, such objectives remain challenging to optimize due to non-differentiable components. 
Recently, Brown~\etal\cite{brown2020smoothap} importantly showcase their effectiveness for image retrieval.
Indeed, when a task aims for a global ranking (\eg, retrieval or quality assessment), it becomes logical to encourage such effect during training.
Inspired by their success, we propose to enhance image quality assessment training signals through the introduction of listwise comparisons.
We derive differentiable regularizers that encapsulate correlation scores between predicted outputs and ground truth mean opinion scores at the mini-batch level.
Complementary to relaxing image content variance constraints, listwise comparisons provide the model with a holistic view of image ranking, which allows model training to adjust prediction scores in a relative manner. 

Our main contributions introduce the concept of content-diverse comparisons when learning image quality assessment models, and can be summarized as:
\begin{enumerate}[topsep=0pt,itemsep=-1ex,partopsep=1ex,parsep=1ex]
\item We relax pairwise constraints to enable the comparison of image pairs with differing content. Pairs can be sampled on-the-fly and within one training dataset. In turn, this better captures wider factors under consideration during human quality assessment.

\item We propose listwise comparisons at a mini-batch level, and derive three differentiable regularizers to encourage correlation with ground-truth image quality rankings.
Such differentiable derivations are applicable to any model architecture without structural changes.
Our regularization terms provide a holistic view during training, which enables the model to adjust image quality scores relative to one another. 
\item We evaluate our proposed learning method on eight different image quality assessment datasets and benchmarks, covering a wide range of unseen distortions and image content. We also show the applicability of our learning method on multiple network architectures and its effectiveness to reach state-of-the-art performance.

\end{enumerate}

\section{Related Work}
\label{sec:related_work}

\paragraph{Image quality assessment} 
assigns a scalar score, used to rank images in terms of their perceived quality.
We consider the full-reference setting~\cite{wang2006modern}, where we have access to the reference image to score the quality of a distorted image. This is particularly useful for evaluating image restoration methods~\cite{liu2013deblurring, ma2017learning, min2019dehazing, zhang2018lpips}.
Other settings can be employed, such as no-reference settings~\cite{wang2006modern} for image aesthetics~\cite{murray2012ava}.  

Traditional metrics for image quality assessment emulate human visual system responses, in order to visualize impairments and distortions~\cite{watson1997digital}, and rely on the assumption that human perceptual quality can be largely explained by the visibility of impairments. Handcrafted metrics attempt to capture well understood visual psychology phenomena such as spatial frequency dependent contrast sensitivity (\eg, SSIM~\cite{wang2004image} and variants). PSNR and SSIM are still commonly used, yet often fall short compared to deep learning models able to capture more complex properties of human visual perception. 
Seminal works propose bespoke convolutional architectures trained from scratch~\cite{kim2017deep,bosse2017wadiqam,prashnani2018pieapp}, whereas Liu~\etal~\cite{liu2017rankiqa} show that fine-tuning networks, pre-trained on ImageNet, are sufficient by following an underlying hypothesis that features extracted from image classification are meaningfully transferable.  Zhang~\etal\cite{zhang2018lpips} go one step further and highlight how to obtain quality scores in the feature space of these pre-trained networks without any fine-tuning.
More recently, Cheon~\etal~\cite{cheon2021iqt} adopt a transformer architecture to predict perceptual quality.
While effective, these works focus solely on characterizing distortions in the image.

In comparison, the broader field of quality of experience considers external factors, \eg, user characteristics or interests~\cite{zhu2016qoe, moller2014quality}, as well as internal factors, \eg, image content~\cite{siahaan2016does,siahaan2016augmenting,siahaan2018semantic}.
This is notably pertinent for videos where every frame comes with its own content and plays a role in the quality score~\cite{li2021unified,liu2018end}.
In a related direction, Ding~\etal\cite{ding2021dists} incorporate texture similarity to the loss for image quality.
We are inspired by this literature and propose to enrich supervision signals. Rather than proposing a new model architecture, we focus on the learning scheme to make comparisons more diverse and more holistic.

\paragraph{Pairwise comparisons} transform the ranking problem into pairwise classification or regression~\cite{liu2011learning, li2014learning}.
This turns the ranking problem into an easier variant, as the model only needs to compare two images at a time.
Liu~\etal\cite{liu2017rankiqa} initially propose a margin-based ranking loss to compare a pair of distorted images with similar content.
Prashnani~\etal\cite{prashnani2018pieapp} learn to regress to the probability of preferring one image over another while Zhang~\etal\cite{zhang2018lpips} simply treat the problem as a binary classification.
Yet, pairwise comparisons are done between two distorted images coming from the same reference image, \ie, similar image content, which does not reflect the true intent of image quality assessment.
Closer to our work, Zhang~\etal~\cite{zhang2021uncertainty} rely on pre-sampled image pairs, across multiple datasets.
We relax current formulations, and enable our model to learn diverse pairwise comparisons of distorted images with differing image content, which are sampled on-the-fly within a single training dataset.

\paragraph{Listwise comparisons} maintain a group structure of ranking, which is ignored in pairwise comparisons~\cite{liu2011learning, li2014learning}.
Indeed, while pairwise comparisons focus on two images at a time, listwise comparisons have a more holistic view of the scores in a mini-batch.
While Cao~\etal~\cite{cao2007learning} study the relevance of learning listwise over pairwise rankings, they don't tie the formulation to any correlation coefficients.
Towards this goal, several works in quality assessment derive a differentiable regularizer for the Pearson coefficient~\cite{li2021unified,li2020norm,liu2018end,ma2018geometric}.
Some attempts have also been made to make Spearman coefficient differentiable through linear programming~\cite{blondel2020fast} or additional layers for sorting~\cite{engilberge2019sodeep}.
While effective, they require fundamental changes and cannot be directly be applicable to any model architectures. We then draw inspiration from the information retrieval literature~\cite{manning2008ir}, where the ranking operation can be approximated through a logistic function~\cite{qin2010general}.
Brown~\etal~\cite{brown2020smoothap} importantly show the positive effect of optimizing such approximated objective for image retrieval.
We build on this literature to derive differentiable listwise regularizers, which ensure holistic properties in a mini-batch and can be applied to any model architecture without any changes.

\section{Method}
\label{sec:method}

\paragraph{Problem formulation.}
During training, we are given a full-reference image quality assessment dataset $\{x^{i}, x^{i}_{\mathrm{ref}}, y^{i} \}_{i=1}^{M}$ of $M$ distorted images $x$ for which a mean opinion score $y$ quantifies the quality with respect to its reference image $x_{\mathrm{ref}}$.
Note that it is common to have multiple distorted images corresponding to the same reference image in image quality assessment datasets. 
The objective is to learn a function $f$ that produces a scalar value $\hat{y}=f(x, x_{\mathrm{ref}})$, in order to predict the mean opinion score and therefore assess the quality of image $x$. In principle, $f$ can be any statistical function and contemporary approaches typically employ a convolutional network.
An effective learning scheme for the scoring function $f$ relies on \emph{pairwise} training~\cite{cao2007learning,zhang2018lpips}. Given two images $x^{i}$ and $x^{j}$ with $i\neq j$, the idea is to learn to predict which image has the better quality. If the mean opinion score $y^{i}$ is higher than $y^{j}$, $\hat{y}^{i}{=}f(x^{i}, x^{i}_{\mathrm{ref}})$ should yield a higher output than $\hat{y}^{j}{=}f(x^{j}, x^{j}_{\mathrm{ref}})$. The model then learns during training to produce a scalar output that quantifies which image in the pair has the better quality. Hence, what matters is to produce a faithful ranking \cf regressing to $y$. Sec.~\ref{sec:method:comp} and~\ref{sec:method:reg} introduce our pair formation strategy and derive our differentiable regularizers, respectively.

\begin{figure}[t]
     \centering
     \hfill
     \subfigure[][\centering \textbf{\textit{Fixed}} pairs with \textbf{\textit{similar}} content]{
     \includegraphics[width=0.25\linewidth]{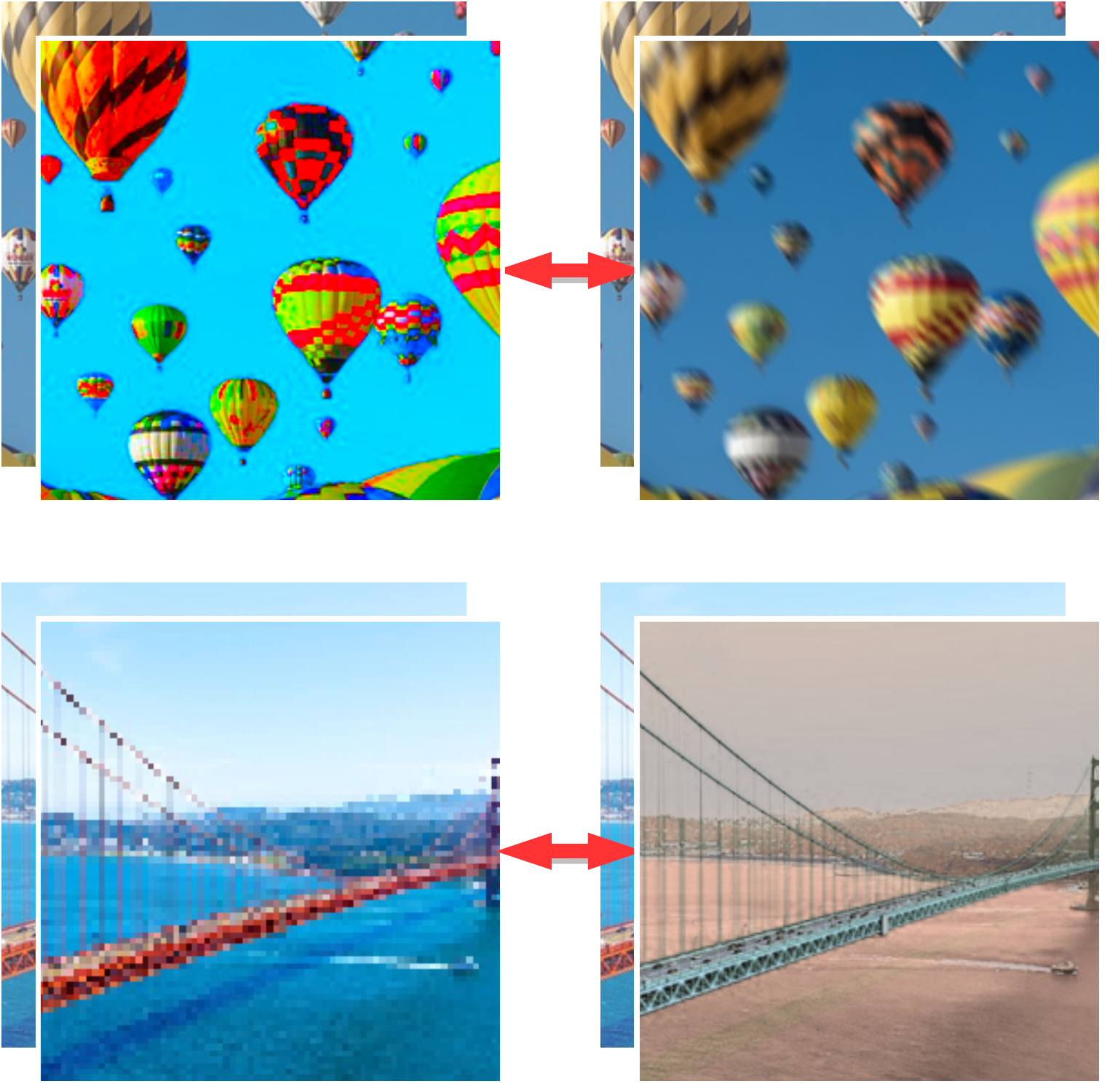}
     \label{fig:two:a}}
     \hfill
     \subfigure[][\centering \textbf{\textit{All}} pairs with \textbf{\textit{similar}} content]{
     \includegraphics[width=0.25\linewidth]{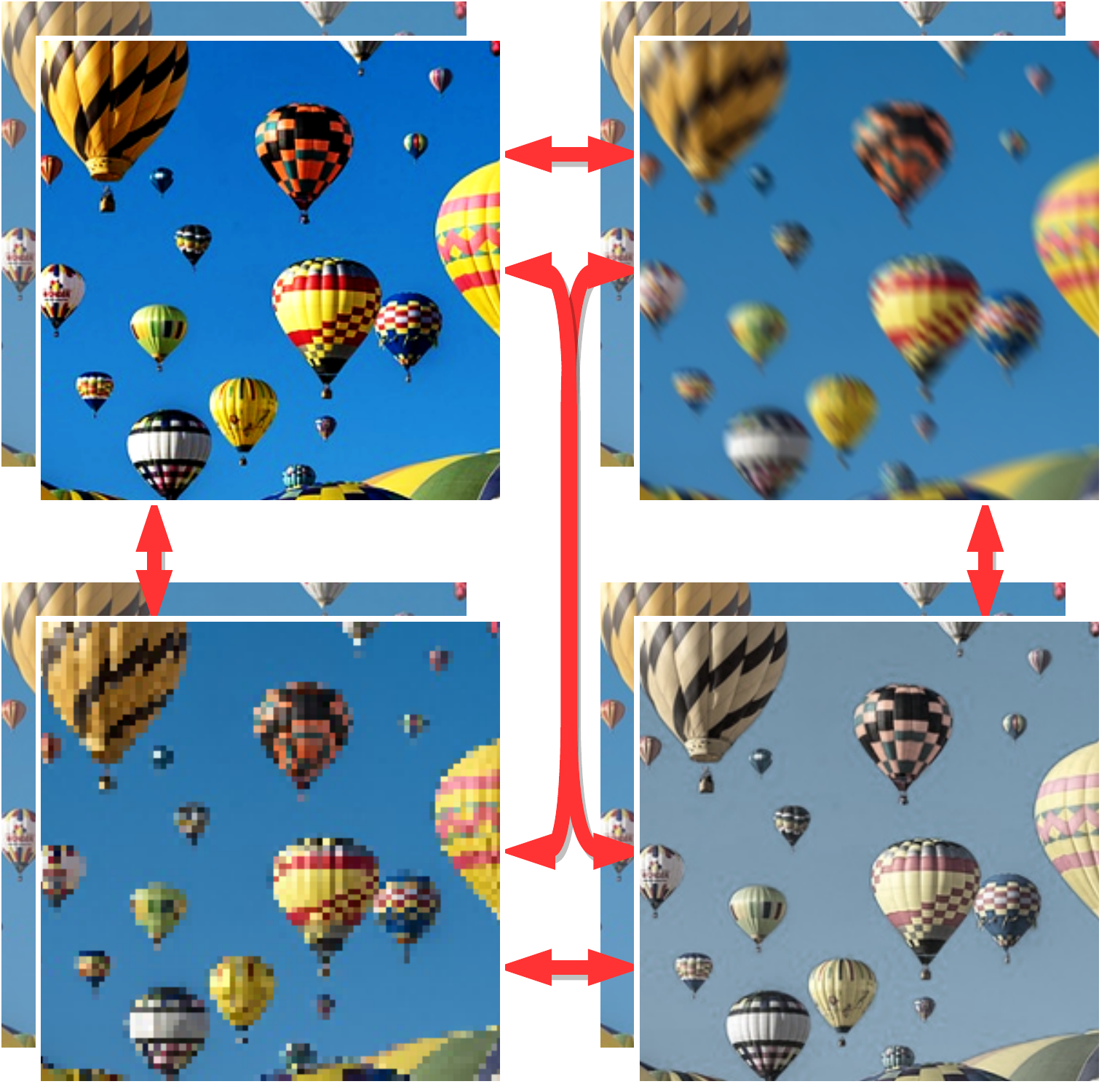}
     \label{fig:two:b}}
     \hfill
     \subfigure[][\centering \textbf{\textit{All}} pairs with \textbf{\textit{differing}} content 
     ]{
     \includegraphics[width=0.25\linewidth]{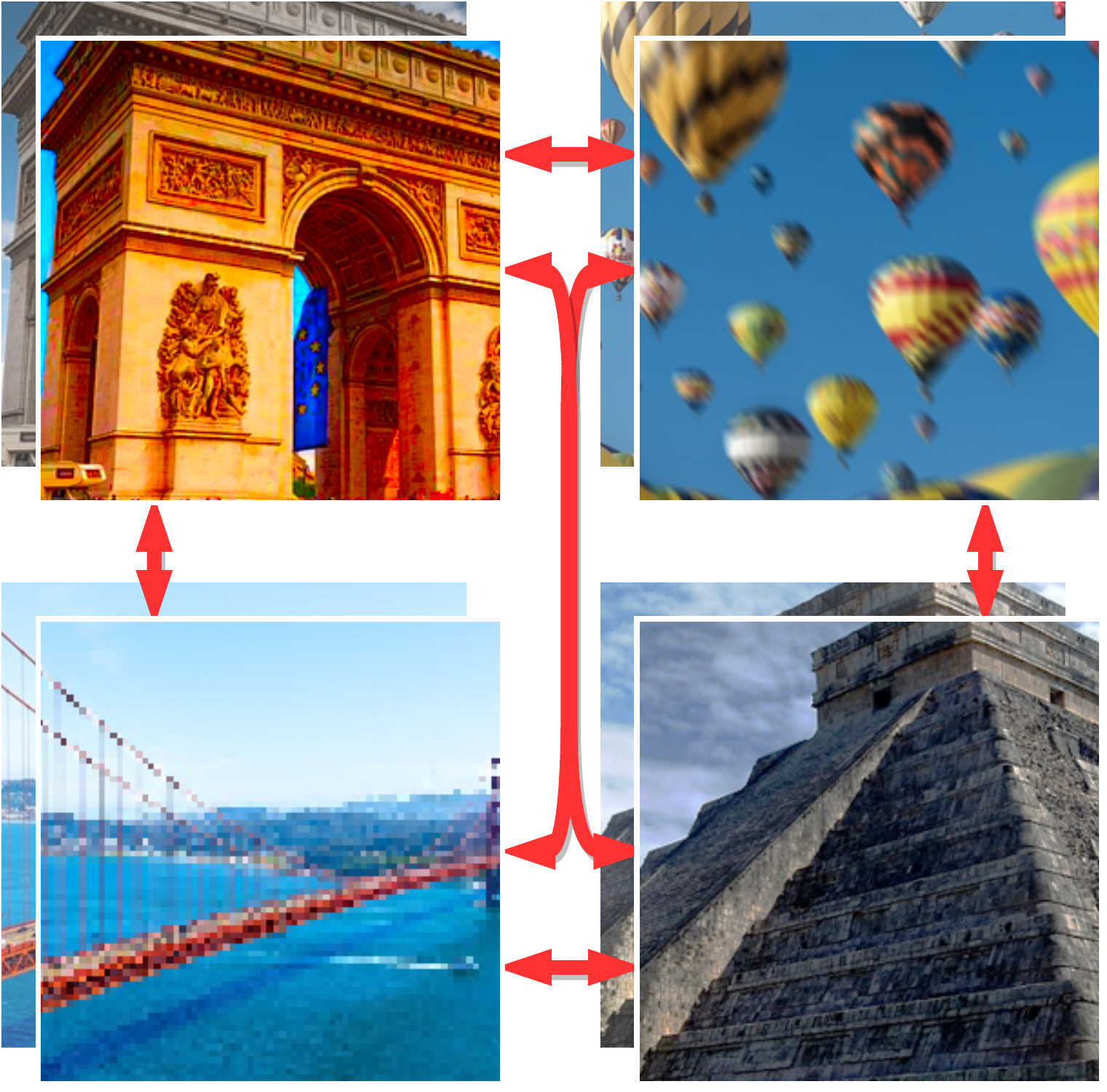}
     \label{fig:two:c}}
     \hfill\null
\caption{\textbf{Image pair formation} in a mini-batch containing four distortions and respective reference images. 
(a) Fixed pairs that contrast only similar content limit comparison count. 
(b) Comparison count can be increased by considering all available distortions for fixed content.
(c) We generalize previous assumptions by unconditionally comparing images with differing image content.
\label{fig:two}}
\vspace{-1em}
\end{figure}

\subsection{Pairwise comparisons with differing content}
\label{sec:method:comp}

\paragraph{Pair formation.}
A standard pair formation scheme~\cite{zhang2018lpips,ding2021dists} is illustrated in Fig.~\ref{fig:two:a}. This first strategy relies on fixed pairs with similar image content, which heavily restricts the number of available pairs. Given $N$ images in a mini-batch, $N/2$ pairs are formed such that every pair compares two distorted images that originate from the \emph{same} reference image.
Fig.~\ref{fig:two:b} illustrates an alternative approach to increase the number of comparisons, by constructing mini-batches comprising a single image content (same reference image), and comparing \emph{all} image pairs ~\cite{liu2017rankiqa}.
Indeed, $(N^2 - N)/2$ pairs can be formed from $N$ images in a mini-batch. 
However, this imposes a strong constraint on the size of the mini-batch during training. Let $D$ be the number of distortions available for a given reference image. Then, under this regime, the size of the mini-batch $N$ cannot be larger than $D$, \ie, $N\leq D$.
Fig.~\ref{fig:two:c} depicts our proposal, where we consider \emph{all} possible pairs yet allow image content to \emph{differ} within a pair. This relaxation allows for  a broader and more diverse definition of valid pairwise comparisons; a generalization of previous schemes.
This content-diverse formulation no longer imposes a constraint on the mini-batch size. More importantly, it enables models to better learn the 
explored intent of the image quality assessment task without any constraint: images should be rankable outwith hard content restrictions. 

\paragraph{Pairwise classification.} 
We learn the scoring function $f$ through pairwise classification. At every iteration, we present to the model a pair of images $x^{i}$ and $x^{j}$ with $i\neq j$, along with their respective reference images. If $x^{i}$ has a better quality than $x^{j}$, this entails that the mean opinion score is higher $y^{i}>y^{j}$. Thus, $f$ should learn to predict a similar ranking $\hat{y}^{i}>\hat{y}^{j}$.
To learn this pairwise classification, we derive a probabilistic model of $y^{i}>y^{j}$ through the Bradley-Terry sigmoid~\cite{bradley1952rank}:

\begin{equation}
p(y^{i}>y^{j})= {1}/\left(1 + \exp \big( - (\hat{y}^{i} - \hat{y}^{j}) / T \big)\right),
\end{equation}
where $T$ is a parametric temperature.
The Bradley-Terry model applies a sigmoid function to the difference of predicted scores $\hat{y}^{i} - \hat{y}^{j}$, which results in a probabilistic model of $y^{i}>y^{j}$.
With this formulation, we can treat the pairwise ranking task as a binary classification.
We then rely on the binary cross-entropy for the loss function:
\begin{equation}
\mathcal{L}_c = \mathbbm{1}[y^i > y^j] \cdot \log(p(y^{i}>y^{j})) + 
\mathbbm{1}[y^i < y^j] \cdot \log(1 - p(y^{i}>y^{j})),
\label{eq:lossC}
\end{equation}
where $\mathbbm{1}[\cdot]$ is the indicator function.
Prashnani~\etal~\cite{prashnani2018pieapp} initially introduced the Bradley-Terry sigmoid for pairwise comparisons, however their loss formulation learns to regress to the ground truth probability $y^{i}>y^{j}$, by minimizing the mean squared error. Thus, the training dataset must include the pairwise probability of preferring $x^{i}$ over $x^{j}$, which requires a different labeling as common datasets usually only assign a mean opinion score. Our binary classification formulation removes this constraint and enables the model to accommodate arbitrary measurement scales, for the scoring ground truth image quality.

\subsection{Holistic listwise comparisons}
\label{sec:method:reg}

\begin{figure}[t]
\begin{minipage}{0.45\textwidth}
\subfigure[][\centering Pairwise scores]{
\begin{overpic}[height=80px, trim=0px 0px -100px 12px,clip]{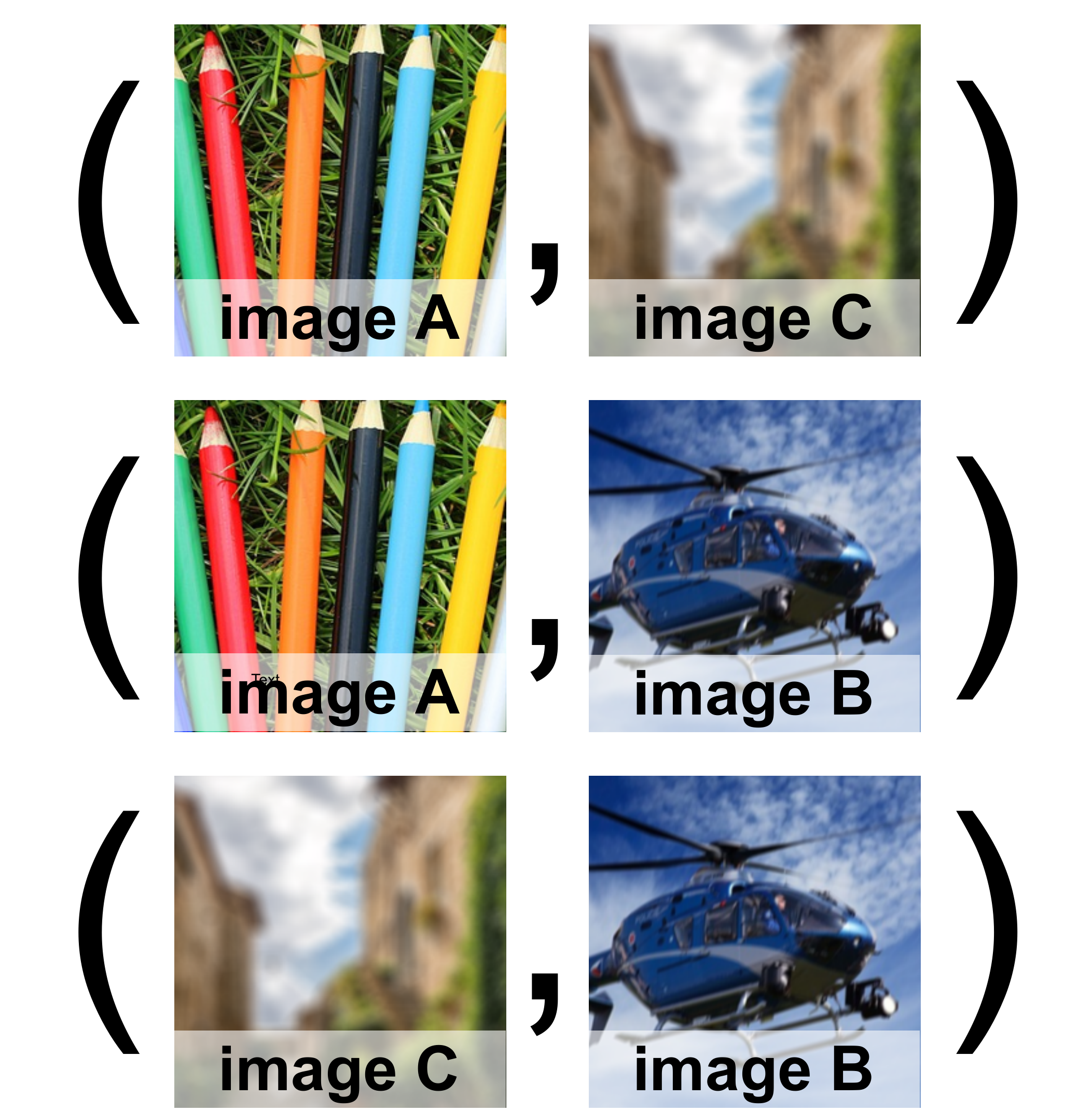}

\put(19,87){\footnotesize \textit{Best}}
\put(48,87){\footnotesize \textit{Worst}}

\put(83,66){\textcolor{teal}{\Large\ding{51}}}
\put(83,38){\textcolor{teal}{\Large\ding{51}}}
\put(83,10){\textcolor{purple}{\Large\ding{55}}}

\end{overpic}
}
\subfigure[][\centering Listwise scores]{
\begin{overpic}[height=80px, trim=0px 0px -50px 0px]{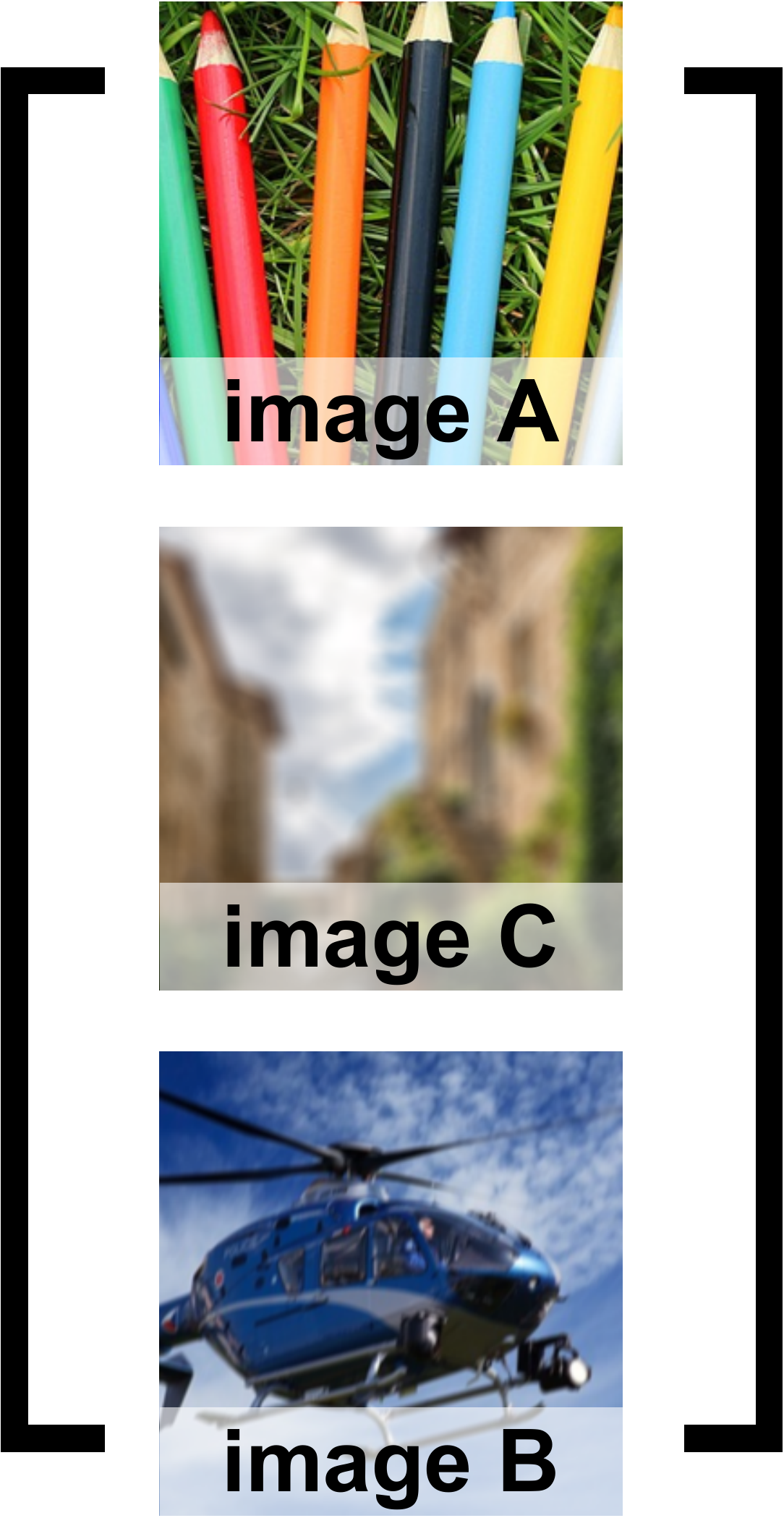}

\put(59,81){{\Large=}}
\put(57,44){\textcolor{purple}{\Large\ding{116}}}
\put(57,10){\textcolor{teal}{\Large\ding{115}}}

\end{overpic}
}
\end{minipage}
\hfill
\begin{minipage}{0.52\textwidth}

\caption{
\textbf{Image ranking.} Three images $A$, $B$ and $C$ with decreasing image quality.
(a) Under pairwise comparisons if the model makes one mistake, \eg rating $C$ as higher quality than $B$, scores cannot be adjusted relative to $A$.
(b) Listwise comparisons enable a holistic view at the mini-batch level. For a similar mistake; richer feedback enables image scores to be adjusted in a relative manner.
}
\label{fig:regularizers}
\end{minipage}
\vspace{-1em}
\end{figure}
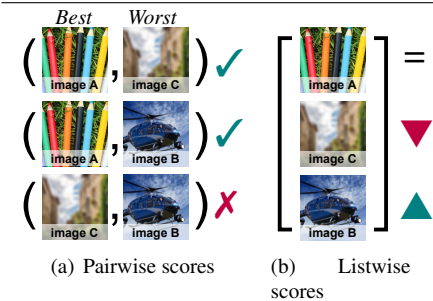

\paragraph{Listwise comparisons.}
Pairwise comparisons only act as a proxy task for image quality assessment. Indeed, during training, the model has a limited view since it only ever compares two distorted images.
Consider Fig.~\ref{fig:regularizers} with a mini-batch of three images $A$, $B$ and $C$. If the model makes a mistake ranking $B$ and $C$, it will be unable to 
dependably adjust the scores relative to $A$, as the two pairwise comparisons involving $A$ are correct. As a result, the loss might over-adjust scores, which may damage the correct comparisons.
We need a constraint to maintain the correct scores and ensure incorrect scores are not over-adjusted.

We propose to compute listwise comparisons through correlation coefficients. Let $\hat{Y} = \{\hat{y}^{1}, \ldots, \hat{y}^{L}\}$ be the predicted scores output by $f$ and $Y{=}\{y^{1}, \ldots, y^{L}\}$ be the related ground truth mean opinion scores for a set of images of size $L$.
Correlation coefficients measure the correlation between $\hat{Y}$ and $Y$. Such measurements provide a holistic view during training as they encourage the model to predict scores relative to each other at the mini-batch level.
We therefore derive regularizers for correlation coefficients at the mini-batch level to better mimic the true objective of image quality assessment.

\paragraph{Correlation coefficients.}
Listwise comparisons introduce helpful holistic properties on the predicted scores. For example, we can ensure a linearity of the scores to avoid producing outlier values; or preserve the ranking to avoid any over-adjustments. To achieve this, we rely on the Pearson coefficient~\cite{pearson1895vii} for linearity, and both Spearman~\cite{spearman1904proof} and Kendall~\cite{kendall1938new} coefficients for ranking.
We introduce regularizers to measure correlation coefficients at the mini-batch level during training: \mbox{$R{=}\lVert 1 - \mathrm{corr}(Y, \hat{Y})\rVert_p$}, where $\lVert \cdot\rVert_p$ is the $p$-norm and both $Y$ and $\hat{Y}$ are of size $N$ (\ie, mini-batch size). By minimizing regularizers in the loss function, the model learns to maximize the correlation between $Y$ and $\hat{Y}$.
While the Pearson coefficient is differentiable, the other two are not. We describe how to derive a regularizer for the Pearson coefficient and then differentiable versions for the Spearman and Kendall coefficients:

\textbf{Pearson linear} correlation coefficient measures the linear relationship between $Y$ and $\hat{Y}$, by computing the ratio between their covariance and the product of their standard deviations: $r(Y,\hat{Y})={{\cov (Y,\hat{Y})}/{\sigma_{Y}\sigma_{\hat{Y}}}}.$
As the Pearson coefficient does not possess any non-differentiable components, and following previous works~\cite{li2021unified,li2020norm,liu2018end,ma2018geometric}, we can directly derive the regularizer: $R_r=\lVert 1 - r(Y, \hat{Y})\rVert_p$.

\textbf{Spearman rank} correlation coefficient measures the linear relationship between the ranking of $Y$ and $\hat{Y}$, by computing the Pearson coefficient between their rank values:
$\rho(Y,\hat{Y}) =\rho(\rank_{Y},\rank_{\hat{Y}})={ {\cov (\rank _{Y},\rank_{\hat{Y}})}/{\sigma_{\rank_{Y}}\sigma_{\rank_{\hat{Y}}}}}$,
where $\rank_{
{Y}} = 1 + \sum_{i\neq j} \mathbbm{1}[(Y_i - Y_j) < 0]$,
and similarly for $\rank_{\hat{Y}}$.
The indicator function in the ranking function makes the Spearman coefficient non-differentiable.
Following works in image retrieval~\cite{brown2020smoothap,qin2010general}, we approximate the indicator function with a temperature-based sigmoid:
\begin{equation}
\mathbbm{1}[(Y^i - Y^j) < 0] \approx 1 / \left( 1+\exp(-(Y^i - Y^j)/T) \right),
\end{equation}
which creates a differentiable Spearman rank $\widetilde{\rho}(\cdot, \cdot)$. We can then derive a corresponding regularizer: \mbox{$R_\rho=\lVert 1 - \widetilde{\rho}(Y, \hat{Y})\rVert_p$}.

\textbf{Kendall rank} correlation coefficient measures the ordinal association between $Y$ and $\hat{Y}$, by computing the number of concordant pairwise comparisons:
$\tau(Y,\hat{Y}) ={\frac {2}{n(n-1)}}\sum_{i<j}\sign(Y^{i}-Y^{j})\sign(\hat{Y}^{i}-\hat{Y}^{j}),$
where $\sign$ is the sign function responsible for making the Kendall coefficient non-differentiable.
Similar to the indicator function, we approximate $\sign$  with a logistic curve. We propose a temperature-based hyperbolic tangent:
\begin{equation}
\sign(Y^{i}-Y^{j}) \approx \tanh((Y^{i}-Y^{j})/T),
\end{equation}
which creates a differentiable Kendall rank $\widetilde{\tau}(\cdot, \cdot)$. We can derive a corresponding regularizer: $R_\tau{=}\lVert 1 - \widetilde{\tau}(Y, \hat{Y})\rVert_p$.
The final loss function then becomes:
\begin{equation}
\mathcal{L} = \mathcal{L}_c + \lambda (R_r+R_\rho+R_\tau),
\label{eq:loss}
\end{equation}
where $\lambda$ controls the contribution of the linearly combined regularizers.

\section{Experiments}
\subsection{Datasets and Evaluation}
\label{sec:data_and_eval}

\paragraph{Setup.}
Unless specified otherwise, we rely on DISTS~\cite{ding2021dists} as the scoring function $f$, which builds on VGG16~\cite{simonyan2015vgg}, pre-trained on ImageNet~\cite{russakovsky2015imagenet} with $L2$ pooling~\cite{henaff2016l2pooling} to extract features at several levels without aliasing.
We train $f$ by minimizing the loss function in Eq.~\ref{eq:loss} with the Adam~\cite{kingma2015adam} optimizer and cosine annealing~\cite{loshchilov2017sgdr}, and set the hyper-parameters as follows: learning rate of $1{\times}10^{-4}$, batch size of $64$, temperature $T$ of $0.01$. For regularization, we use the absolute value (\ie, $p = 1$) and set $\lambda$ to $1.0$.
During training, we randomly crop $256{\times}256$ image patches and apply random rotations ($\{0^{\circ}, 90^{\circ}, 180^{\circ}\}$) with horizontal flipping. During test, the model is evaluated using full images in all experiments.

\paragraph{Datasets.}

Following previous works~\cite{ding2021dists,cheon2021iqt}, we train our model on the KADID-10k~\cite{lin2019kadid} dataset, which contains $25$ different distortions at five different levels of severity for a total of 10,125 distorted images. Distortion types include traditional artifacts, such as those from blurring or compression, but also in-the-wild artifacts, such as denoising or non-linear brightness artifacts.
To evaluate generalization capabilities, we compare on three traditional image quality assessment datasets with broad distortion types:
LIVE~\cite{sheikh2003live}, CSIQ \cite{larson2010csiq} and TID2013~\cite{ponomarenko2015tid}. They contain 779, 866 and 3000 distorted images, respectively. We additionally evaluate on five other benchmarks in the appendix.

\paragraph{Evaluation.}
We report the Pearson linear (PLCC)~\cite{pearson1895vii}, Spearman rank (SRCC) \cite{spearman1904proof} and Kendall rank (KRCC)~\cite{kendall1938new} correlation coefficients.
When computing the PLCC, we follow Ding~\etal~\cite{ding2021dists} and fit a four-parameter logistic regression (4LP) function:
$\hat{y}_{\textrm{4lp}} = (\eta_1 - \eta_2) / (1 + \exp(-(\hat{y} - \eta_3)/\eta_4)) + \eta_2$, where $\{\eta_i\}_{i=1}^4$ are estimated through least squares.

\begin{table*}[t]
\centering
\tablestyle{2pt}{1.1}
\resizebox{0.75\width}{!}{
\begin{tabular}{l|l|ccc|ccc|ccc}
\multirow{2}{*}{Pair formation} & \multirow{2}{*}{Content}
& \multicolumn{3}{c|}{LIVE\cite{sheikh2003live}}
& \multicolumn{3}{c|}{CSIQ\cite{larson2010csiq}}
& \multicolumn{3}{c}{TID2013\cite{ponomarenko2015tid}} \\
 & & PLCC & SRCC & KRCC & 
 PLCC & SRCC & KRCC &
 PLCC & SRCC & KRCC \\
\shline
fixed pairs & similar & 0.958 & 0.964 & 0.828 & \textbf{0.951} & 0.953 & 0.807 & 0.889 & 0.878 & 0.690 \\
all pairs &  similar  & 0.961 & 0.966 & 0.836 & 0.948 & 0.952 & 0.805 & 0.902 & 0.892 & 0.709 \\
all pairs & differing & \textbf{0.963} & \textbf{0.968} & \textbf{0.842} & 0.950 & \textbf{0.954} & \textbf{0.809} & \textbf{0.908} & \textbf{0.897} & \textbf{0.717} \\
\end{tabular}
}
\caption{\textbf{Pair formation} for pairwise comparisons. We compare a \textit{fixed} pair formation with \textit{all} available pairs in a mini-batch. Pairs with no constraints on image content improve upon more constrained pair formations.
\label{tab:pair}}
\vspace{-1em}
\end{table*}

\label{sec:experiments}

\subsection{Ablations}
\label{sec:experiments:ablation}

\paragraph{Pair formation.}
We report results of the investigated strategies in Tab.~\ref{tab:pair}. 
Liu~\etal~\cite{liu2017rankiqa} previously note that having an order of magnitude more comparative pairs, per image, yields improvement in the correlation metrics. We corroborate this experimentally and observe that, controlling for the number of images provided, the model learns to correctly rank more image pairs on average using \emph{all pairs} compared with \emph{fixed pairs}.
By further removing the image content restriction and allowing comparisons between \mbox{\emph{differing}} image content, our proposal leads to a further improvement. In contrast with Liu~\etal~\cite{liu2017rankiqa}, our proposal removes the limitation regarding mini-batch size. For example, if a training dataset contains a variable number of distortions per reference image, our method can gracefully handle such data without additional problem (\cf \emph{all pairs} with \emph{same} content).
Our approach enables models to learn from comparisons exhibiting highly diverse content, which better resembles the underdetermined problem of real-world image quality assessment. 

\begin{table}[t]
\begin{minipage}{0.6\textwidth}
\tablestyle{2pt}{1.1}
\resizebox{0.75\width}{!}{
\begin{tabular}{ccc|ccc|ccc|ccc}
\multicolumn{3}{c|}{\scriptsize Regularizer} 
& \multicolumn{3}{c|}{LIVE\cite{sheikh2003live}}
& \multicolumn{3}{c|}{CSIQ\cite{larson2010csiq}}
& \multicolumn{3}{c}{TID2013\cite{ponomarenko2015tid}} \\
$R_r$ & $R_\rho$ & $R_\tau$ & PLCC & SRCC & KRCC 
& PLCC & SRCC & KRCC & PLCC & SRCC & KRCC \\
\shline
& & & 0.963 & 0.968 & 0.842 & 0.950 & 0.954 & 0.809 & 0.908 & 0.897 & 0.717 \\
\hline
\ding{51} & & & 0.962 & 0.967 & 0.839 & 0.952 & 0.956 & 0.812 & 0.906 & 0.896 & 0.715 \\
& \ding{51} & & 0.960 & 0.966 & 0.835 & 0.953 & 0.957 & 0.815 & 0.910 & 0.901 & 0.723 \\
& & \ding{51} & 0.962 & 0.968 & 0.840 & 0.950 & 0.955 & 0.811 & 0.910 & 0.900 & 0.722 \\
\hline
\ding{51} & \ding{51} & & 0.960 & 0.966 & 0.837 & 0.954 & 0.959 & 0.819 & 0.908 & 0.899 & 0.718 \\
& \ding{51} & \ding{51} & 0.961 & 0.967 & 0.838 & 0.941 & 0.960 & 0.821 & 0.909 & 0.900 & 0.721 \\
\ding{51} & & \ding{51} & 0.960 & 0.966 & 0.837 & 0.954 & 0.959 & 0.820 & 0.912 & 0.903 & 0.725 \\
\hline
\ding{51} & \ding{51} & \ding{51} & \textbf{0.964} & \textbf{0.969} & \textbf{0.843} & \textbf{0.957} & \textbf{0.960} & \textbf{0.824} & \textbf{0.915} & \textbf{0.907} & \textbf{0.731} \\
\end{tabular}
}
\end{minipage}
\hfill
\begin{minipage}{0.38\textwidth}
\caption{
\textbf{Regularization} on $r$, $\rho$ and $\tau$.
Multiple regularizers encourage linear properties and rank preservation, which consistently improve performance, see text for further detail.
\label{tab:reg}
}
\end{minipage}
\vspace{-1em}
\end{table}

\paragraph{Regularizers.}
Previous work has focused mainly on pairwise image comparisons (\eg~\cite{zhang2018lpips,ding2021dists}), and also explored a Pearson regularizer (\eg,~\cite{li2021unified,li2020norm,liu2018end,ma2018geometric}). Yet, none have assessed the interaction of multiple regularizers with complementary ranking properties.
Consequently, we propose to provide models with a richer understanding of image quality, by encouraging the learning of a global ranking. Our regularization signals enable learning of listwise rankings at the mini-batch level, towards a holistic understanding of image quality.

Tab.~\ref{tab:reg} evaluates the effect of the regularizers at an individual level, as well as in combination.
Regularizers can be considered to possess complementary components which we evaluate. The Pearson regularization $R_r$ ensures linearity while the Spearman regularization $R_\rho$ preserves the ranking of the predicted scores in the mini-batch. The Kendall regularization $R_\tau$ shares the same spirit with the classification loss function $\mathcal{L}_c$ (Eq.~\ref{eq:lossC}), as it ensures the pairwise ranking of all available pairs in the mini-batch, and not only the magnitude of the predicted score.
Overall, these different regularizers on top of the diverse pair formation provide complementary properties for the model to learn a linear global score for ranking images.
In isolation, and also when combined with one other regularizer, we observe that $R_r$ has a negligible effect. This suggests that $\mathcal{L}_c$ already produces linear scores.
Both $R_\rho$ and $R_\tau$ share a broadly common goal and focus on the ranking, which is complementary to $\mathcal{L}_c$. Individually, we observe these two terms tend to provide similar benefit while combining both together may result in some redundancy.
Interestingly, it is when all regularizers are included that the performance is best. In this setting, $R_r$ does yield a positive effect, across all three considered benchmarks.
Deriving multiple regularizers with complementary roles enhances the type of comparisons learned during training.

A limitation of the listwise regularizers concerns the batch size.  When going from 64 to 8, we observe a drop in performance by up to 5.06\% on TID2013. We provide more details in the supplementary materials (Sec.~\ref{sec:supp:batch}). Thus, a lower batch size reduces the effectiveness of listwise comparisons.

\paragraph{Scoring functions.}
Our contributions are not tied to a specific quality scoring function $f$ (\ie, model architecture). We evidence this claim by evaluating alternative full-reference image quality scoring functions, trained with our proposed learning strategy:
(i) LPIPS~\cite{zhang2018lpips} takes the difference between the distorted image and its reference at several levels in VGG16 \cite{simonyan2015vgg} to produce a scalar predicted score;
(ii) LPIPS with the addition of an antialiasing module. As image quality assessment requires to capture small differences in the image space and such terms are reported to be beneficial~\cite{henaff2016l2pooling};
(iii) DISTS~\cite{ding2021dists} differs by learning to weight the contribution of every feature level.
Scoring functions, each employing our training contributions (Sec.~\ref{sec:method}), are evaluated in Tab.~\ref{tab:arch}.
We observe that our training scheme benefits all the selected models, as we improve upon their original scores on the benchmarks.
This gain originates from comparison learning instead of training directly to regress mean opinion scores. The latter lacks a mechanism to induce score linearity and ordering, which our regularizers address.
Additionally, we evaluate the effect of the patch size in the supplementary materials (Sec.~\ref{sec:supp:git}) as LPIPS and DISTS have been trained with different patch sizes.
We select DISTS for our following experiments (Sec.~\ref{sec:experiments:sota}) as it yields the best overall performance. We however highlight that our proposed training strategy is applicable to multiple state-of-the-art image quality scoring functions.

\begin{table}[t]
\begin{minipage}{0.58\textwidth}
\tablestyle{2pt}{1.1}
\resizebox{0.81\width}{!}{
\begin{tabular}{l|ccc|ccc|ccc}
\multirow{2}{*}{Model}  
& \multicolumn{3}{c|}{LIVE\cite{sheikh2003live}}
& \multicolumn{3}{c|}{CSIQ\cite{larson2010csiq}}
& \multicolumn{3}{c}{TID2013\cite{ponomarenko2015tid}} \\
 & PLCC & SRCC & KRCC & PLCC & SRCC & KRCC & PLCC & SRCC & KRCC \\
\shline
 LPIPS & 0.959 & 0.966 & 0.832 & 0.952 & 0.958 & 0.815 & 0.895 & 0.887 & 0.700 \\
 +$L2$ pool & 0.959 & 0.964 & 0.828 & 0.945 & \textbf{0.964} & \textbf{0.827} & 0.892 & 0.886 & 0.699 \\
 DISTS & \textbf{0.964} & \textbf{0.969} & \textbf{0.843} & \textbf{0.957} & 0.960 & 0.824 & \textbf{0.915} & \textbf{0.907} & \textbf{0.731} \\
\end{tabular}
}
\end{minipage}
\hfill
\begin{minipage}{0.4\textwidth}
\caption{
\textbf{Scoring functions} for perceptual similarity. Our learning strategy can be applied to different architectures without structural modifications, see text for further detail. 
\label{tab:arch}
}
\end{minipage}
\vspace{-1em}
\end{table}

\subsection{State-of-the-art comparisons}
\label{sec:experiments:sota}

\begin{table}[t]
\begin{minipage}{0.58\textwidth}
\tablestyle{2pt}{1.1}
\resizebox{0.75\width}{!}{
\begin{tabular}{l|ccc|ccc|ccc}
\multirow{2}{*}{Method} & \multicolumn{3}{c|}{LIVE\cite{sheikh2003live}} &  \multicolumn{3}{c|}{CSIQ\cite{larson2010csiq}} & \multicolumn{3}{c}{TID2013\cite{ponomarenko2015tid}} \\
 & PLCC & SRCC & KRCC & PLCC & SRCC & KRCC & PLCC & SRCC & KRCC \\
\shline
PSNR & 0.865 & 0.873 & 0.680  & 0.819 & 0.810 & 0.601  & 0.677 & 0.687 & 0.496 \\
SSIM~\cite{wang2004image} & 0.937 &0.948 & 0.796 & 0.852 & 0.865 & 0.680 & 0.777 & 0.727 & 0.545 \\
MS-SSIM~\cite{wang2003msssim} &  0.940 & 0.951 & 0.805  &  0.889 & 0.906 & 0.730 & 0.830 & 0.786 &  0.605 \\
VSI~\cite{zhang2014vsi}  & 0.948 & 0.952 & 0.806 & 0.928 & 0.942 & 0.786  & \textbf{0.900} & 0.897 & \textbf{0.718} \\
MAD~\cite{larson2010csiq} & \textbf{0.968} & 0.967 & 0.842 & \textbf{0.950} & \textbf{0.947} & 0.797 & 0.827 & 0.781 & 0.604\\
VIF~\cite{sheikh2006vif}  & 0.960 & 0.964 & 0.828 & 0.913 & 0.911 & 0.743 &  0.771   & 0.677 &  0.518 \\
FSIMc~\cite{zhang2011fsim} & 0.961 & 0.965 & 0.836 &	0.919 & 0.931 &	0.769  & 0.877 & 0.851 & 0.667 \\
NLPD~\cite{laparra2016nlpd} & 0.932 & 0.937 & 0.778 & 0.923 & 0.932 & 0.769 & 0.839 &  0.800  & 0.625 \\
GMSD~\cite{xue2013gmsd}  & 0.957 & 0.960 & 0.827 & 0.945  & 0.950 & \textbf{0.804} & 0.855 & 0.804 & 0.634 \\
\hline
WaDIQaM~\cite{bosse2017wadiqam} & 0.940 & 0.947 & 0.791 & 0.901 & 0.909 & 0.732 & 0.834 & 0.831  & 0.631 \\
PieAPP~\cite{prashnani2018pieapp} & 0.908 & 0.919 & 0.750 & 0.877 & 0.892 & 0.715 & 0.859 & 0.876 & 0.683 \\
LPIPS~\cite{zhang2018lpips} & 0.934 & 0.932 & 0.765 & 0.896 & 0.876 & 0.689 & 0.749 & 0.670 & 0.497 \\
DISTS~\cite{ding2021dists} & 0.954 & 0.954 & 0.811 & 0.928 & 0.929 & 0.767& 0.855 & 0.830 & 0.639 \\
IQT~\cite{cheon2021iqt} & -- & \textbf{0.970} & \textbf{0.849} & -- & 0.943 & 0.799 & -- & \textbf{0.899} & 0.717 \\
\hline
\textit{Ours} & \textbf{0.964} & \textbf{0.969} & \textbf{0.843} & \textbf{0.957} & \textbf{0.960} & \textbf{0.824} & \textbf{0.915} & \textbf{0.907} & \textbf{0.731} \\
\end{tabular}
}
\end{minipage}
\hfill
\begin{minipage}{0.4\textwidth}

\caption{\textbf{Comparison on traditional distortions}
with top-2 results highlighted in bold.
Only IQT and our method achieve a consistent performance across all three datasets.
Our method based on the DISTS architecture improves upon the original training, confirming the benefits our proposed training scheme.
\label{tab:sota:iqa}}
\end{minipage}
\vspace{-1em}
\end{table}

\paragraph{Quantitative evaluation.}
Tab.~\ref{tab:sota:iqa} compares approaches on LIVE~\cite{sheikh2003live}, CSIQ~\cite{larson2010csiq} and TID 2013~\cite{ponomarenko2015tid}.
Ideally, a generic image quality assessment method should yield high scores, regardless of the type of distortions and image content present in the data.
In practice, we observe that methods tend to struggle to achieve good overall performance. For example, MAD correlates well with mean opinion scores on the LIVE and CSIQ datasets, yet struggles on TID2013 where there is a large gap with VSI. Disparity in dataset collection protocols (\eg, controlled \vs uncontrolled environments) may introduce additional dataset annotation variance and challenges.  
Interestingly, similar behaviour can be observed on several deep learning based methods (\eg, LPIPS and DISTS), which under-perform on TID2013. 
Only our method and the recent IQT~\cite{cheon2021iqt} are able to perform consistently across all three datasets. IQT proposes a transformer, different from convolutional networks most commonly used in contemporary image quality assessment, which led the authors to win the NTIRE'21 challenge~\cite{gu2021ntire}.
Still, our method achieves the best overall scores, which evidences generalization ability across a variety of datasets.
Furthermore, we observe that our method,
instantiated here using the DISTS architecture, outperforms their original training scheme.
In principle, we expect similar gains when applying our training scheme to alternative transformer-based architectures (\eg, IQT), which we consider a promising direction for future work.
In the supplementary materials (Sec.~\ref{sec:supp:quantitative}), we provide six additional benchmarks where our perceptual loss also yields state-of-the-art results.

\newcommand{\imgWidthSR}{0.18}
\newcommand{\hspaceSizeSR}{0.001}
\newcommand{\imageNumSR}{002} 
\newcommand{\imageSetSR}{14} %

\newcommand{\subfigSRone}{bicubic}
\newcommand{\subfigSRtwo}{psnr}
\newcommand{\subfigSRthree}{official}
\newcommand{\subfigSRfour}{ours}
\newcommand{\subfigSRfive}{HR}

\newcommand{\trimLeft}{0}
\newcommand{\trimBottom}{0}
\newcommand{\trimRight}{0}
\newcommand{\trimTop}{0}
 
\begin{figure*}[t]
\centering
\subfigure[\tiny Bicubic interpolation]{\includegraphics[trim=\trimLeft px \trimBottom px \trimRight px \trimTop px, clip, width=\imgWidthSR\textwidth]{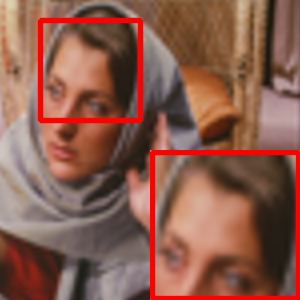}
\label{fig:sr_example:LR}}
\hfill
\subfigure[\tiny $L2$ loss]{\includegraphics[trim=\trimLeft px \trimBottom px \trimRight px \trimTop px, clip, width=\imgWidthSR\textwidth]{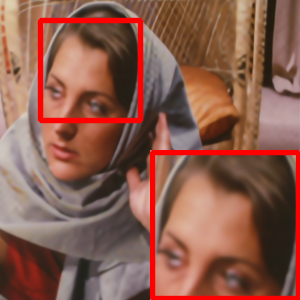}
\label{fig:sr_example:psnr}}  
\hfill
\subfigure[\tiny $L2$ + perceptual loss~\cite{Johnson2016Perceptual}]{\includegraphics[trim=\trimLeft px \trimBottom px \trimRight px \trimTop px, clip, width=\imgWidthSR\textwidth]{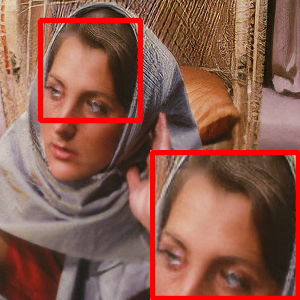}
\label{fig:sr_example:official}} 
\hfill
\subfigure[\tiny $L2$ + perceptual loss (\textit{ours})]{\includegraphics[trim=\trimLeft px \trimBottom px \trimRight px \trimTop px, clip, width=\imgWidthSR\textwidth]{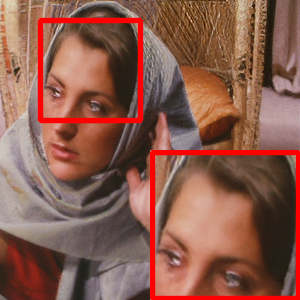}
\label{fig:sr_example:ours}}
\hfill
\subfigure[\tiny ground truth]{\includegraphics[trim=\trimLeft px \trimBottom px \trimRight px \trimTop px, clip, width=\imgWidthSR\textwidth]{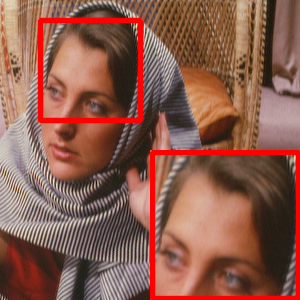}
\label{fig:sr_example:HR}}
\caption{\textbf{Qualitative application} of our image quality metric as a training objective for ${\times}4$ super-resolution.
We evaluate different objectives for ESRGAN~\cite{wang2018esrgan} (b-d). All improve over the blurriness of bicubic interpolation (a). Ours (d) reduces subtle over-sharpening artifacts present in with Johnson~\etal\cite{Johnson2016Perceptual} (c), best viewed in color with digital zoom.
}
\label{fig:sr_example}
\vspace{-1em}
\end{figure*}

\paragraph{Qualitative application.}
While image quality assessment primarily targets the scoring of distortions arising from image restoration methods, it may alternatively be used to train the respective restoration models.
We consider the super-resolution task, and select ESRGAN~\cite{wang2018esrgan} which includes a PSNR-oriented $L2$ loss, that optimizes
image reconstruction, and a further perceptual loss function component.
Following Wang~\etal\cite{wang2018esrgan}, we train the models on the DIV2K~\cite{Agustsson2017div2k} dataset.
Fig.~\ref{fig:sr_example} provides qualitative 
results for a ${\times}4$ SR task on a sample from Set14~\cite{zeyde2010single}.
While optimizing for the $L2$ loss reduces the blurriness of a simple bicubic interpolation, it still misses high-frequency textures (Fig.~\ref{fig:sr_example:LR} \vs Fig.~\ref{fig:sr_example:psnr}).
Adding a perceptual component to the loss function helps. 
In lieu of the Johnson~\etal\cite{Johnson2016Perceptual}
perceptual loss, in the original ESRGAN~\cite{wang2018esrgan} formulation,
we substitute our model as the perceptual component. 
We observe that over-sharpening artifacts are reduced when our model is used (Fig.~\ref{fig:sr_example:official} \vs Fig.~\ref{fig:sr_example:ours}).
In the supplementary materials (Sec.~\ref{sec:supp:qualitative_sr}), we provide additional examples where our perceptual loss also benefits for reducing the presence of other artifacts.
Our brief exploration illustrates the potential of our approach as an objective for downstream tasks where perceptual quality is of prime importance.

\section{Conclusion}
\label{sec:conclusion}

We present a learning scheme with content-diverse comparisons for learning an image quality assessment model.
Firstly, we formulate comparisons in a manner that relaxes image content constraints. This makes pairwise comparisons content-diverse and better leverages the available ground truth annotations.
Second, we include listwise comparisons to provide a holistic view to the training signal. %
We derive differentiable regularizers from correlation coefficients to ensure the linearity and rankings of predicted scores at the mini-batch level, which enables models to adjust scores in a relative manner.
Finally, we validate the effectiveness of our learning scheme, where we report state-of-the-art performance and explore the potential of the trained image quality model for downstream imaging tasks.

\clearpage

\bibliography{egbib}

\clearpage

\appendix

\section{Supplementary Material}
\label{sec:supp}\noindent

\setcounter{equation}{0}
\setcounter{figure}{0}
\setcounter{table}{0}
\setcounter{page}{1}
\makeatletter
\renewcommand{\theequation}{S\arabic{equation}}
\renewcommand{\thefigure}{S\arabic{figure}}
\renewcommand{\thetable}{S\arabic{table}}

We provide additional materials to supplement our main paper. 
Sec.~\ref{sec:supp:ablations} provides additional ablation studies on the image patch size (Sec.~\ref{sec:supp:git}), the batch size used training (Sec.~\ref{sec:supp:batch}).
Sec.~\ref{sec:supp:quantitative} evaluates on six additional benchmarks where we also achieve state-of-the-art results.
In Sec.~\ref{sec:supp:qualitative_sr} we provide additional qualitative examples for the application of our trained model as a training objective for the super-resolution task.
Sec.~\ref{sec:supp:licenses} reports the licenses of the datasets used in our experimental work.
Finally, Sec.~\ref{sec:supp:societal} discusses potential societal impacts and limitations of our paper.

\subsection{Ablations}
\label{sec:supp:ablations}
\subsubsection{Training image patch size}
\label{sec:supp:git}

We assess the effect of the patch size used during training. For example, LPIPS \cite{zhang2018lpips} relies on a patch size of $64{\times}64$ while DISTS~\cite{ding2021dists} considers $256{\times}256$.
Tab.~\ref{tab:supp:patch} shows that the larger the patch size, the better the performance.
This concurs with our assumption that image content matters. A larger patch size captures more image content whereas a smaller one only provides a local view.
Increasing the patch size beyond $256{\times}256$ creates memory issues, and might not be optimal given that it reduces the variety of samples seen during training.
Following Ding~\etal\cite{ding2021dists}, we then adopt a patch size of $256{\times}256$ in our experiments.

\begin{table}[h!]
\centering
\tablestyle{1.5pt}{1.1}
\begin{tabular}{c|ccc|ccc|ccc}
Patch
& \multicolumn{3}{c|}{LIVE\cite{sheikh2003live}}
& \multicolumn{3}{c|}{CSIQ\cite{larson2010csiq}}
& \multicolumn{3}{c}{TID2013\cite{ponomarenko2015tid}} \\
size & PLCC & SRCC & KRCC & PLCC & SRCC & KRCC & PLCC & SRCC & KRCC \\
\shline
64 & 0.936 & 0.945 & 0.792 & 0.943 & 0.948 & 0.795 & 0.823 & 0.796 & 0.605 \\
128 & 0.949 & 0.959 & 0.820 & 0.949 & 0.955 & 0.810 & 0.879 & 0.870 & 0.681 \\
256 & \textbf{0.964} & \textbf{0.969} & \textbf{0.843} & \textbf{0.957} & \textbf{0.960} & \textbf{0.824} & \textbf{0.915} & \textbf{0.907} & \textbf{0.731} \\
\end{tabular}
\vspace{0.25em}
\caption{\textbf{Image patch size} used for training. A larger patch size captures more image content and results in improved performance.
\label{tab:supp:patch}}
\end{table}

\subsubsection{Training batch size}
\label{sec:supp:batch}
We assess the effect of the batch size during training. For the experiment, we vary the batch size, and keep the number of epochs similar.
Tab.~\ref{tab:supp:batch} shows that the bigger the batch size, the better the performance.
A bigger batch size increases the number 
of comparisons that are made at every training iteration. Thus, the benefit of all pairwise and listwise comparisons becomes much more effective.
Increasing the batch size above 64 creates memory issues (with current hardware). As such, we adopt a batch size of 64 in our experiments.

\begin{table*}[t]
\centering
\tablestyle{2.5pt}{1.1}

\begin{tabular}{c|ccc|ccc|ccc}
Batch
& \multicolumn{3}{c|}{LIVE\cite{sheikh2003live}}
& \multicolumn{3}{c|}{CSIQ\cite{larson2010csiq}}
& \multicolumn{3}{c}{TID2013\cite{ponomarenko2015tid}} \\
size &  
 PLCC & SRCC & KRCC & 
 PLCC & SRCC & KRCC &
 PLCC & SRCC & KRCC \\
\shline
8  & 0.949 & 0.955 & 0.811 & 0.944 & 0.952 & 0.804 & 0.887 & 0.880 & 0.694 \\
16  & 0.953 & 0.959 & 0.821 & 0.944 & 0.952 & 0.803 & 0.899 & 0.893 & 0.711 \\
32  & 0.957 & 0.964 & 0.831 & 0.952 & 0.956 & 0.813 & 0.909 & 0.902 & 0.722 \\
64 & \textbf{0.964} & \textbf{0.969} & \textbf{0.843} & \textbf{0.957} & \textbf{0.960} & \textbf{0.824} & \textbf{0.915} & \textbf{0.907} & \textbf{0.731} \\
\end{tabular}
\vspace{0.25em}
\caption{\textbf{Batch size} used for training over 10 epochs. A bigger batch size increases the the number of comparisons, which results in improved performance.
\label{tab:supp:batch}}
\end{table*}

\subsection{Quantitative evaluation}
\label{sec:supp:quantitative}

\paragraph{Setup.}
First, We assess the ability to rank image pairs in a two-alternative forced choice (2AFC) setting, using the BAPPS~\cite{zhang2018lpips} dataset of 26,904 image pairs.
Second, we compare on four datasets with task-specific artifacts arising from image restoration and enhancement algorithms: Liu13~\cite{liu2013deblurring} for deblurring, Ma17~\cite{ma2017learning} for super-resolution, Min19~\cite{min2019dehazing} for dehazing, and Tian18~\cite{tian2018rendering} for rendering. They contain 1200, 1620, 600 and 140 distorted images, respectively.
Third, we evaluate on the PIPAL dataset~\cite{gu2020pipal}, which was used in the recent NTIRE'21 and '22 challenges~\cite{gu2021ntire}.

For the 2AFC experiment, we follow Zhang~\etal~\cite{zhang2018lpips} and report the 2AFC score 
\mbox{$qp+(1-q)(1-p)$},
where $q$ is the ground truth probability and $p$ is the model prediction.

\subsubsection{Pairwise comparisons}
Tab.~\ref{tab:sota:bapps} compares our method using the BAPPS~\cite{zhang2018lpips} dataset in a two-alternative forced choice (2AFC) setting.
Given two $64{\times}64$ image patches with different distortions, yet originating from the same reference image, the 2AFC task consists of correctly predicting which patch has the best quality.
Following Ding~\etal\cite{ding2021dists}, we report results on the test set of distortions, that originate from real algorithms for colorization, video deblurring, frame interpolation and super-resolution.
As LPIPS has been trained on a separate set of BAPPS, it provides a strong baseline.
Even though this evaluation relies on comparing images with similar content, our method yields the best overall performance; 
despite being trained on the broader task of comparing differing image content.

\begin{table*}
\centering
\begin{minipage}{0.5\linewidth}
\tablestyle{2pt}{1.1}
\resizebox{0.9\width}{!}{
\begin{tabular}{l|ccccc}
\multirow{2}{*}{Method} & \multirow{2}{*}{\shortstack{Colori-\\zation}} & \multirow{2}{*}{\shortstack{Video\\deblur.}} & \multirow{2}{*}{\shortstack{Frame\\interp.}} & \multirow{2}{*}{\shortstack{Super-\\resolution}} & All \\
 & \\
\shline
Human & 0.688 & 0.671 & 0.686 & 0.734 & 0.695 \\
\hline
PSNR                          & 0.624 & 0.590 & 0.543 & 0.642 & 0.614 \\
SSIM~\cite{wang2004image}     & 0.522 & 0.583 & 0.548 & 0.613 & 0.617 \\
MS-SSIM~\cite{wang2003msssim} & 0.522 & 0.589 & 0.572 & 0.638 & 0.596 \\
VSI~\cite{zhang2014vsi}       & 0.597 & 0.591 & 0.568 & 0.668 & 0.622 \\
MAD~\cite{larson2010csiq}     & 0.490 & 0.593 & 0.581 & 0.655 & 0.599 \\
VIF~\cite{sheikh2006vif}      & 0.515 & 0.594 & 0.597 & 0.651 & 0.603 \\
FSIMc~\cite{zhang2011fsim}    & 0.573 & 0.590 & 0.581 & 0.660 & 0.615 \\
NLPD~\cite{laparra2016nlpd}   & 0.528 & 0.584 & 0.552 & 0.655 & 0.600 \\
GMSD~\cite{xue2013gmsd}       & 0.517 & 0.594 & 0.575 & 0.676 & 0.613 \\
\hline
PieAPP~\cite{prashnani2018pieapp} & 0.594 & 0.582 & 0.598 & 0.685 & 0.626 \\
LPIPS~\cite{zhang2018lpips} & 0.625 & \textbf{0.605} & \textbf{0.630} & 0.705 & \textbf{0.641} \\
DISTS~\cite{ding2021dists} & \textbf{0.627} & 0.600 & 0.625 & \textbf{0.710} & \textbf{0.641} \\
\hline
\textit{Ours} & \textbf{0.632} & \textbf{0.605} & \textbf{0.631} & \textbf{0.712} & \textbf{0.645}
\\
\end{tabular}
}
\end{minipage}
\hfill
\begin{minipage}{0.45\linewidth}
\caption{\textbf{Comparison on image pair ranking} in a 2AFC setting on the BAPPS\cite{zhang2018lpips} dataset, with top-2 results highlighted in bold.
While we train with diverse-content and holistic properties, our method is on par with LPIPS trained specifically on this task.
\label{tab:sota:bapps}}
\end{minipage}
\vspace{-1em}
\end{table*}

\subsubsection{Task-specific distortions}
Tab.~\ref{tab:sota:restoration} further compares our approach on distortions arising in specific tasks: Liu13~\cite{liu2013deblurring} on deblurring, Ma17~\cite{ma2017learning} on super-resolution, Min19~\cite{min2019dehazing} on dehazing, and Tian18~\cite{tian2018rendering} on rendering.
In other words, each dataset is task-specific as they only include distortions originating from methods related to the respective tasks. For example, Liu13 provides mean opinion scores for images pertaining to five different deblurring methods.
A similar behavior to the broad distortions results may be observed here; consistent performance across all datasets is challenging.
For example, the performance suddenly drops for LPIPS and PieApp methods on Tian18 rendering images. They tend to struggle with the domain gap present in this dataset and cannot quantify distortions, regardless of the image domain.
In this context, only DISTS and our method can provide consistent and high correlation scores on all four datasets.

\begin{table*}[t]
\centering
\tablestyle{2pt}{1.1}
\resizebox{0.9\width}{!}{
\begin{tabular}{l|ccc|ccc|ccc|ccc}
\multirow{2}{*}{Method} & \multicolumn{3}{c|}{Liu13\cite{liu2013deblurring} (deblurring)} &  \multicolumn{3}{c|}{Ma17\cite{ma2017learning} (SR)} & \multicolumn{3}{c|}{Min19\cite{min2019dehazing} (dehazing)} & \multicolumn{3}{c}{Tian18\cite{tian2018rendering} (rendering)} \\
 & PLCC & SRCC & KRCC & PLCC & SRCC & KRCC & PLCC & SRCC & KRCC & PLCC & SRCC & KRCC \\
\shline
PSNR & 0.807 & 0.803 & 0.599 & 0.611 & 0.592 & 0.414 & 0.754 & 0.740 & 0.555 & 0.605 & 0.536 & 0.377 \\
SSIM~\cite{wang2004image} & 0.763 & 0.777 & 0.574 & 0.654 & 0.624 & 0.440 & 0.715 & 0.692 & 0.513 & 0.420 & 0.230 & 0.156 \\
MS-SSIM~\cite{wang2003msssim} & 0.899 & 0.898 & 0.714 & 0.815 & 0.795 & 0.598 & 0.699 & 0.687 & 0.503 & 0.386 & 0.396 & 0.264 \\
VSI~\cite{zhang2014vsi} & 0.919 & 0.920 & 0.745 & 0.736 & 0.710 & 0.514 & 0.730 & 0.696 & 0.511 & 0.512 & 0.531 & 0.363 \\
MAD~\cite{larson2010csiq} & 0.901 & 0.897 & 0.714 & 0.873 & 0.864 & 0.669 & 0.543 & 0.605 & 0.437 & 0.690 & 0.622 & 0.441 \\
VIF~\cite{sheikh2006vif} & 0.879 & 0.864 & 0.672 & 0.849 & 0.831 & 0.638 & 0.740 & 0.667 & 0.504 & 0.429 & 0.259 & 0.173 \\
FSIMc~\cite{zhang2011fsim} & 0.923 & 0.921 & 0.749 & 0.769 & 0.747 & 0.548 & 0.747 & 0.695 & 0.515 & 0.496 & 0.476 & 0.324 \\
NLPD~\cite{laparra2016nlpd} & 0.862 & 0.853 & 0.657 & 0.749 & 0.732 & 0.535 & 0.616 & 0.608 & 0.442 & 0.594 & 0.463 & 0.316 \\
GMSD~\cite{xue2013gmsd} & 0.927 & 0.918 & 0.746 & 0.861 & 0.851 & 0.661 & 0.675 & 0.663 & 0.489 & 0.631 & 0.479 & 0.329 \\
\hline
PieAPP~\cite{prashnani2018pieapp} & 0.752 & 0.786 & 0.583 & 0.791 & 0.771 & 0.591 & 0.749 & 0.725 & 0.547 & 0.352 & 0.298 & 0.207 \\
LPIPS~\cite{zhang2018lpips} & 0.853 & 0.867 & 0.675 & 0.809 & 0.788 & 0.687 & \textbf{0.825} & 0.777 & 0.592 & 0.387 & 0.311 & 0.213 \\
DISTS~\cite{ding2021dists} & \textbf{0.940} & \textbf{0.941} & \textbf{0.784} & \textbf{0.887} & \textbf{0.878} & \textbf{0.697} & 0.816 & \textbf{0.789} & \textbf{0.600} & \textbf{0.694} & \textbf{0.671} & \textbf{0.485} \\
\hline
\textit{Ours} & \textbf{0.937} & \textbf{0.941} & \textbf{0.786} & \textbf{0.898} & \textbf{0.883} & \textbf{0.700} & \textbf{0.819} & \textbf{0.801} & \textbf{0.607} & \textbf{0.720} & \textbf{0.662} & \textbf{0.482} \\
\end{tabular}
} 
\caption{\textbf{Comparison on task-specific distortions} with top-2 results highlighted in bold.
While learning-based method generalize well across datasets, they struggle when there is a domain gap as in image rendering. DISTS and our method achieve a consistent performance.
\label{tab:sota:restoration}}
\vspace{-1em}
\end{table*}

\subsubsection{GAN distortions}
\label{sec:supp:pipal}

We provide additional quantitative comparisons with LPIPS~\cite{zhang2018lpips} and DISTS~\cite{ding2021dists} on the training split of PIPAL~\cite{gu2020pipal}. The NTIRE'21 challenge~\cite{gu2021ntire} relied on the PIPAL dataset as it provides more recent and challenging distortions coming for example from GAN-based restotation methods. As the ground truth labels are not available for the testing split, we instead rely on the training split for evaluation.
Tab.~\ref{tab:supp:pipal} shows that our training scheme outperforms the original DISTS on this challenging dataset.
Note that none of the models evaluated in Tab.~\ref{tab:supp:pipal} have been trained on any splits of PIPAL (\ie, this is a similar setting to experiments in Tab.~\ref{tab:sota:iqa} or Tab.~\ref{tab:sota:restoration}).

\begin{table*}[h!]
\centering
\tablestyle{6pt}{1.1}

\begin{tabular}{c|ccc}
Method
& \multicolumn{3}{c}{PIPAL train split\cite{gu2020pipal}}\\
 &
 PLCC & SRCC & KRCC\\
\shline
LPIPS~\cite{zhang2018lpips} & 0.611 & 0.573 & 0.405\\
DISTS~\cite{ding2021dists} & 0.592 & 0.579 & 0.408 \\
\hline
\textit{Ours} & \textbf{0.701} & \textbf{0.671} & \textbf{0.484}\\
\end{tabular}
\vspace{0.25em}
\caption{\textbf{Comparison on more challenging distortions} from the PIPAL dataset. Our training scheme improves upon the original DISTS and LPIPS in all metrics.
\label{tab:supp:pipal}}
\end{table*}

\input{sec/supp_fig.tex}

\subsection{Qualitative application results}
\label{sec:supp:qualitative_sr}

We provide additional qualitative results when using our trained model as a training objective for a super-resolution application.
Similar to the main paper, we select images from Set14~\cite{zeyde2010single} while ERSGAN models are trained on DIV2K~\cite{Agustsson2017div2k}.
Figure~\ref{fig:supp:sr_example:14:013} shows other benefits of our model: we notably observe a better enhancement of shapes (\eg, letters or zebra stripes) and the absence of an unwanted color cast, that may appear in grayscale images when employing an alternative perceptual loss.
This additional brief exploration further illustrates the potential to use our approach as an objective for downstream tasks, such as super-resolution, where perceptual quality is of prime importance.

\subsection{Dataset licenses and implementation}
\label{sec:supp:licenses}

We rely on nine image quality assessment datasets: one for training and eight for evaluation.
No new dataset has been collected in this paper, and datasets we rely on for our experiments have been previously published:
Kadid-10k~\cite{lin2019kadid} mentions the Pixabay license; other datasets (LIVE~\cite{sheikh2003live}, CSIQ~\cite{larson2010csiq}, TID~\cite{ponomarenko2015tid}, BAPPS~\cite{zhang2018lpips}, Liu13~\cite{liu2013deblurring}, Ma17~\cite{ma2017learning}, Min19~\cite{min2019dehazing}, Tian18~\cite{tian2018rendering}) provide a copyright for research usage (\ie, non-commercial purposes). Future work will aim to provide implementation under popular learning frameworks \eg~\cite{paszke2019pytorch,abadi2016tensorflow,MindSpore}. 

\subsection{Societal impact and limitations}
\label{sec:supp:societal}
Automating image quality assessment may induce the eradication of related manual tasks, that currently constitute valid paid work for humans.
However we concur with Bhardwaj~\etal~\cite{bhardwaj2020unsupervised}; human verification is likely to continue to be required in both the short and medium term. Given the difficulty of image quality assessment, human verification remains the gold standard.

Learning-based perceptual image metrics typically rely on manually-rated images as training data. Such data include human rater variance, and may reflect the biases of human annotators. Reducing or removing the dependency on human labels provides one direction towards mitigating this particular form of bias. 

Finally, as highlighted in the paper, image content matters in image quality assessment. An important question could be to assess whether some contents create a systematic source of bias either from the annotations (\ie, data bias) or the learned metrics (\ie, model bias).

\end{document}

%% file: sec/supp_fig.tex
\renewcommand{\imgWidthSR}{0.15}
\renewcommand{\hspaceSizeSR}{-1.45} 

\renewcommand{\subfigSRone}{bicubic}
\renewcommand{\subfigSRtwo}{psnr}
\renewcommand{\subfigSRthree}{official}
\renewcommand{\subfigSRfour}{ours}
\renewcommand{\subfigSRfive}{HR}


\renewcommand{\trimLeft}{0}
\renewcommand{\trimBottom}{0}
\renewcommand{\trimRight}{0}
\renewcommand{\trimTop}{0}

\begin{figure*}[!ht]
\centering
\begin{tabular}[b]{c}
\tiny Bicubic interpolation \\ 
\includegraphics[trim=\trimLeft px \trimBottom px \trimRight px \trimTop px, clip, width=\imgWidthSR\textwidth]{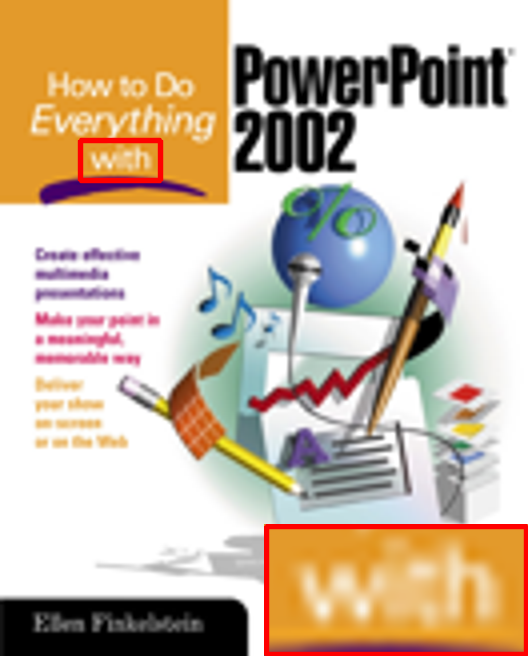}\\[1em]
\includegraphics[trim=\trimLeft px \trimBottom px \trimRight px \trimTop px, clip, width=\imgWidthSR\textwidth]{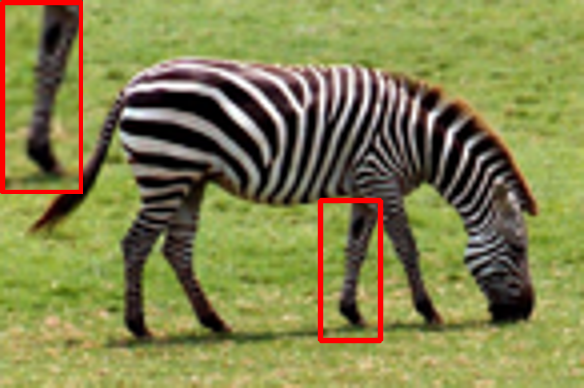}\\[1em]
\includegraphics[trim=\trimLeft px \trimBottom px \trimRight px \trimTop px, clip, width=\imgWidthSR\textwidth]{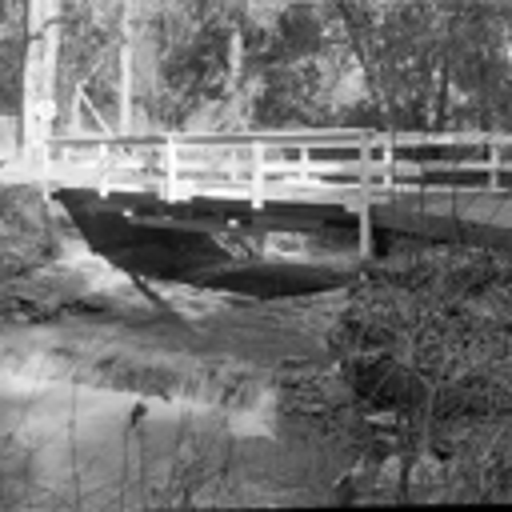}
\end{tabular}
\begin{tabular}[b]{c}
\tiny $L2$ loss \\
\includegraphics[trim=\trimLeft px \trimBottom px \trimRight px \trimTop px, clip, width=\imgWidthSR\textwidth]{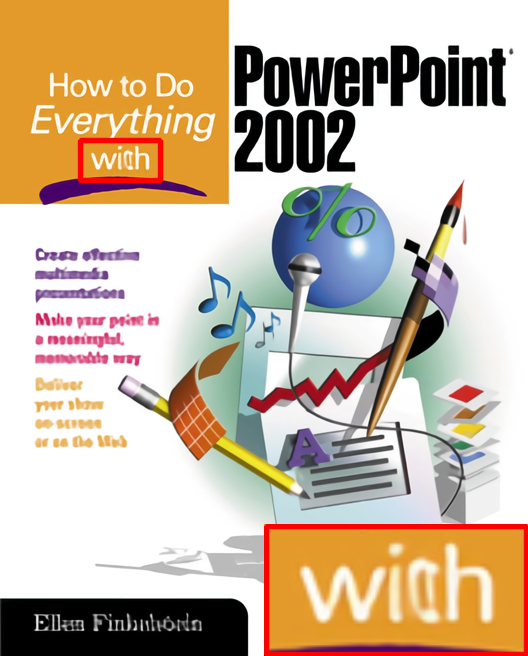}\\[1em]
\includegraphics[trim=\trimLeft px \trimBottom px \trimRight px \trimTop px, clip, width=\imgWidthSR\textwidth]{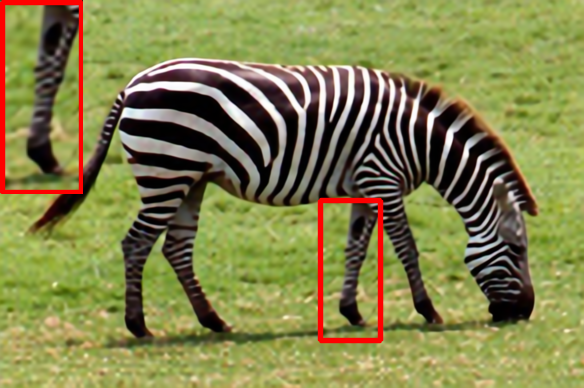}\\[1em]
\includegraphics[trim=\trimLeft px \trimBottom px \trimRight px \trimTop px, clip, width=\imgWidthSR\textwidth]{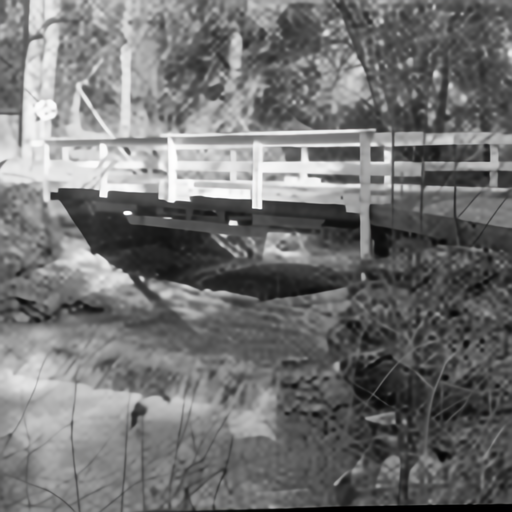}
\end{tabular}
\begin{tabular}[b]{c}
\tiny $L2$ + perceptual loss~\cite{Johnson2016Perceptual} \\
\includegraphics[trim=\trimLeft px \trimBottom px \trimRight px \trimTop px, clip, width=\imgWidthSR\textwidth]{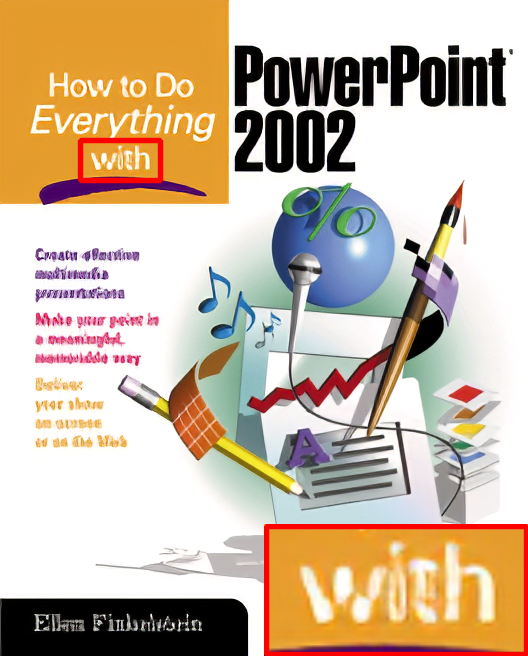}\\[1em]
\includegraphics[trim=\trimLeft px \trimBottom px \trimRight px \trimTop px, clip, width=\imgWidthSR\textwidth]{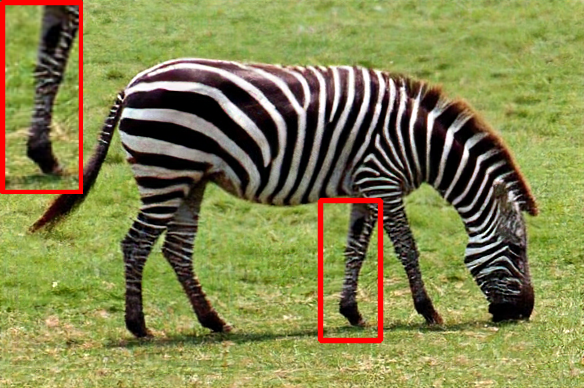}\\[1em]
\includegraphics[trim=\trimLeft px \trimBottom px \trimRight px \trimTop px, clip, width=\imgWidthSR\textwidth]{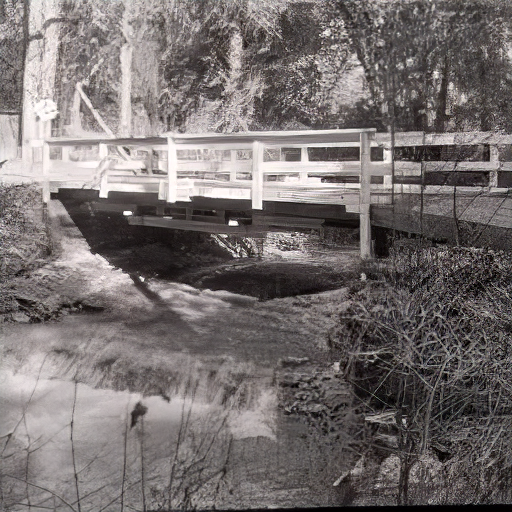}
\end{tabular}
\begin{tabular}[b]{c}
\tiny $L2$ + perceptual loss (\textit{ours}) \\
\includegraphics[trim=\trimLeft px \trimBottom px \trimRight px \trimTop px, clip, width=\imgWidthSR\textwidth]{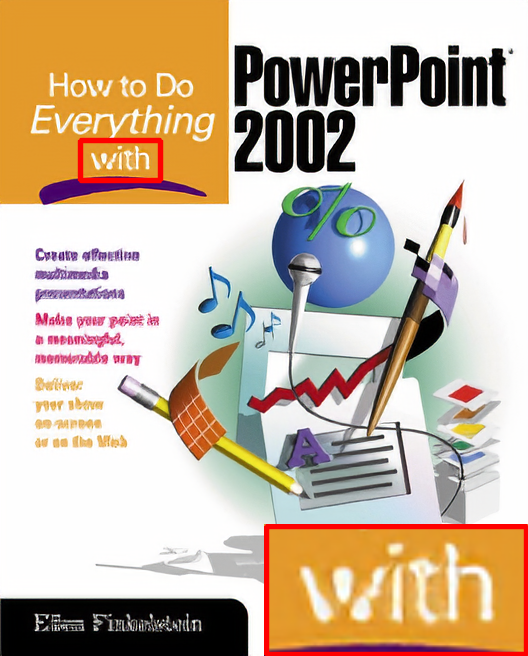}\\[1em]
\includegraphics[trim=\trimLeft px \trimBottom px \trimRight px \trimTop px, clip, width=\imgWidthSR\textwidth]{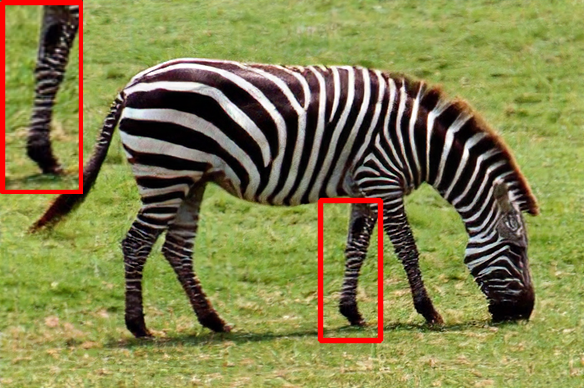}\\[1em]
\includegraphics[trim=\trimLeft px \trimBottom px \trimRight px \trimTop px, clip, width=\imgWidthSR\textwidth]{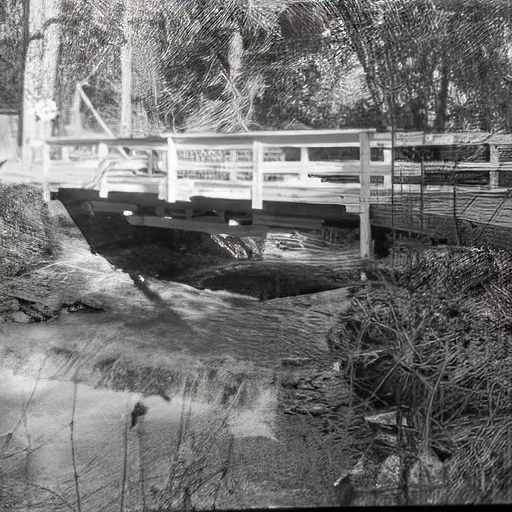}
\end{tabular}
\begin{tabular}[b]{c}
\tiny Ground truth \\
\includegraphics[trim=\trimLeft px \trimBottom px \trimRight px \trimTop px, clip, width=\imgWidthSR\textwidth]{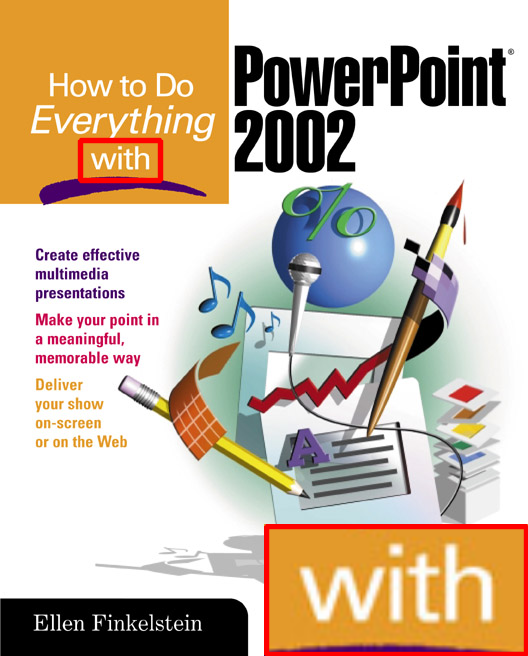}\\[1em]
\includegraphics[trim=\trimLeft px \trimBottom px \trimRight px \trimTop px, clip, width=\imgWidthSR\textwidth]{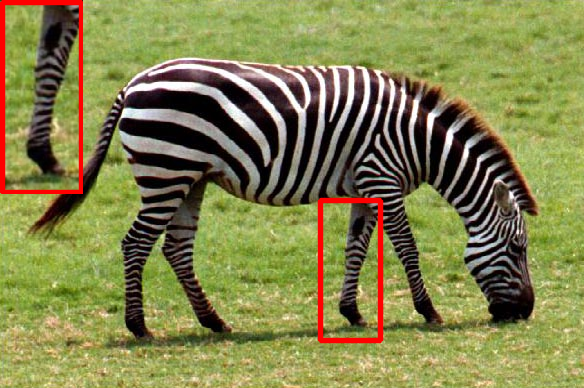}\\[1em]
\includegraphics[trim=\trimLeft px \trimBottom px \trimRight px \trimTop px, clip, width=\imgWidthSR\textwidth]{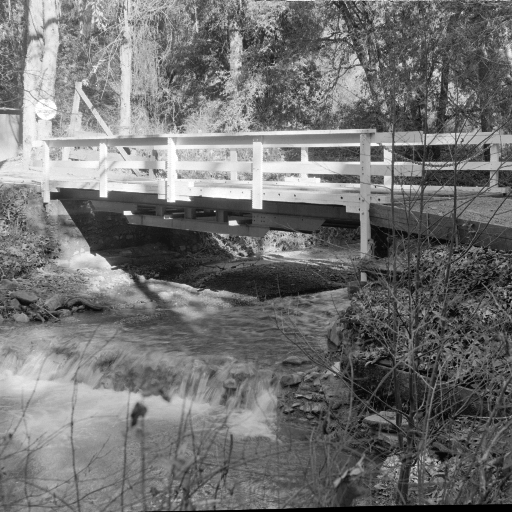}
\end{tabular}
\caption{\textbf{Qualitative application} of our image quality assessment as a training objective for a ${\times}4$ super-resolution task, using Set14~\cite{zeyde2010single} images. Best viewed in color with digital zoom.
We evaluate different training objectives for ESRGAN~\cite{wang2018esrgan} (columns $2 \mbox{--} 4$) and compare with a naive bicubic interpolation (column 1).
\textit{Top row:} While still not perfect, letter shapes appear more realistic with our perceptual loss.
\textit{Middle row:} This is confirmed on the zebra stripes.
\textit{Bottom row:} Moreover, compared with perceptual loss from Johnson~\etal\cite{Johnson2016Perceptual}, we do not exhibit a color cast and can handle a grayscale image.
}
\label{fig:supp:sr_example:14:013}
\end{figure*}

%% file: AIPQ_arxiv.bbl
\begin{thebibliography}{64}
\providecommand{\natexlab}[1]{#1}
\providecommand{\url}[1]{\texttt{#1}}
\expandafter\ifx\csname urlstyle\endcsname\relax
  \providecommand{\doi}[1]{doi: #1}\else
  \providecommand{\doi}{doi: \begingroup \urlstyle{rm}\Url}\fi

\bibitem[Min()]{MindSpore}
Mind{S}pore.
\newblock \url{https://www.mindspore.cn/}.
\newblock Accessed: 2022-10-07.

\bibitem[Abadi et~al.(2016)Abadi, Barham, Chen, Chen, Davis, Dean, Devin,
  Ghemawat, Irving, Isard, et~al.]{abadi2016tensorflow}
Mart{\'\i}n Abadi, Paul Barham, Jianmin Chen, Zhifeng Chen, Andy Davis, Jeffrey
  Dean, Matthieu Devin, Sanjay Ghemawat, Geoffrey Irving, Michael Isard, et~al.
\newblock {T}ensor{F}low: a system for {L}arge-{S}cale machine learning.
\newblock In \emph{12th USENIX symposium on operating systems design and
  implementation (OSDI 16)}, pages 265--283, 2016.

\bibitem[Agustsson and Timofte(2017)]{Agustsson2017div2k}
Eirikur Agustsson and Radu Timofte.
\newblock {NTIRE} 2017 challenge on single image super-resolution: Dataset and
  study.
\newblock In \emph{CVPRw}, 2017.

\bibitem[Bhardwaj et~al.(2020)Bhardwaj, Fischer, Ball{\'e}, and
  Chinen]{bhardwaj2020unsupervised}
Sangnie Bhardwaj, Ian Fischer, Johannes Ball{\'e}, and Troy Chinen.
\newblock An unsupervised information-theoretic perceptual quality metric.
\newblock \emph{arXiv preprint arXiv:2006.06752}, 2020.

\bibitem[Blondel et~al.(2020)Blondel, Teboul, Berthet, and
  Djolonga]{blondel2020fast}
Mathieu Blondel, Olivier Teboul, Quentin Berthet, and Josip Djolonga.
\newblock Fast differentiable sorting and ranking.
\newblock In \emph{ICML}, 2020.

\bibitem[Bosse et~al.(2017)Bosse, Maniry, M{\"u}ller, Wiegand, and
  Samek]{bosse2017wadiqam}
Sebastian Bosse, Dominique Maniry, Klaus-Robert M{\"u}ller, Thomas Wiegand, and
  Wojciech Samek.
\newblock Deep neural networks for no-reference and full-reference image
  quality assessment.
\newblock \emph{TIP}, 27\penalty0 (1):\penalty0 206--219, 2017.

\bibitem[Bradley and Terry(1952)]{bradley1952rank}
Ralph~Allan Bradley and Milton~E Terry.
\newblock Rank analysis of incomplete block designs: I. the method of paired
  comparisons.
\newblock \emph{Biometrika}, 39\penalty0 (3/4):\penalty0 324--345, 1952.

\bibitem[Brown et~al.(2020)Brown, Xie, Kalogeiton, and
  Zisserman]{brown2020smoothap}
Andrew Brown, Weidi Xie, Vicky Kalogeiton, and Andrew Zisserman.
\newblock Smooth-ap: Smoothing the path towards large-scale image retrieval.
\newblock In \emph{ECCV}, 2020.

\bibitem[Cao et~al.(2007)Cao, Qin, Liu, Tsai, and Li]{cao2007learning}
Zhe Cao, Tao Qin, Tie-Yan Liu, Ming-Feng Tsai, and Hang Li.
\newblock Learning to rank: from pairwise approach to listwise approach.
\newblock In \emph{ICML}, 2007.

\bibitem[Cheon et~al.(2021)Cheon, Yoon, Kang, and Lee]{cheon2021iqt}
Manri Cheon, Sung-Jun Yoon, Byungyeon Kang, and Junwoo Lee.
\newblock Perceptual image quality assessment with transformers.
\newblock In \emph{CVPRw}, 2021.

\bibitem[Ding et~al.(2021)Ding, Ma, Wang, and Simoncelli]{ding2021dists}
Keyan Ding, Kede Ma, Shiqi Wang, and Eero~P Simoncelli.
\newblock Image quality assessment: Unifying structure and texture similarity.
\newblock \emph{PAMI}, 2021.

\bibitem[Engilberge et~al.(2019)Engilberge, Chevallier, P{\'e}rez, and
  Cord]{engilberge2019sodeep}
Martin Engilberge, Louis Chevallier, Patrick P{\'e}rez, and Matthieu Cord.
\newblock Sodeep: a sorting deep net to learn ranking loss surrogates.
\newblock In \emph{CVPR}, 2019.

\bibitem[Gu et~al.(2020)Gu, Cai, Chen, Ye, Ren, and Dong]{gu2020pipal}
Jinjin Gu, Haoming Cai, Haoyu Chen, Xiaoxing Ye, Jimmy~S Ren, and Chao Dong.
\newblock Pipal: a large-scale image quality assessment dataset for perceptual
  image restoration.
\newblock In \emph{ECCV}, 2020.

\bibitem[Gu et~al.(2021)Gu, Cai, Dong, Ren, Qiao, Gu, and Timofte]{gu2021ntire}
Jinjin Gu, Haoming Cai, Chao Dong, Jimmy~S Ren, Yu~Qiao, Shuhang Gu, and Radu
  Timofte.
\newblock {NTIRE} 2021 challenge on perceptual image quality assessment.
\newblock In \emph{CVPRw}, 2021.

\bibitem[H{\'e}naff and Simoncelli(2016)]{henaff2016l2pooling}
Olivier~J H{\'e}naff and Eero~P Simoncelli.
\newblock Geodesics of learned representations.
\newblock In \emph{ICLR}, 2016.

\bibitem[Johnson et~al.(2016)Johnson, Alahi, and
  Fei-Fei]{Johnson2016Perceptual}
Justin Johnson, Alexandre Alahi, and Li~Fei-Fei.
\newblock Perceptual losses for real-time style transfer and super-resolution.
\newblock In \emph{ECCV}, 2016.

\bibitem[Kendall(1938)]{kendall1938new}
Maurice~G Kendall.
\newblock A new measure of rank correlation.
\newblock \emph{Biometrika}, 30\penalty0 (1/2):\penalty0 81--93, 1938.

\bibitem[Kim and Lee(2017)]{kim2017deep}
Jongyoo Kim and Sanghoon Lee.
\newblock Deep learning of human visual sensitivity in image quality assessment
  framework.
\newblock In \emph{CVPR}, 2017.

\bibitem[Kingma and Ba(2015)]{kingma2015adam}
Diederik~P Kingma and Jimmy Ba.
\newblock Adam: A method for stochastic optimization.
\newblock In \emph{ICLR}, 2015.

\bibitem[Laparra et~al.(2016)Laparra, Ball{\'e}, Berardino, and
  Simoncelli]{laparra2016nlpd}
Valero Laparra, Johannes Ball{\'e}, Alexander Berardino, and Eero~P Simoncelli.
\newblock Perceptual image quality assessment using a normalized laplacian
  pyramid.
\newblock \emph{Electronic Imaging}, 2016\penalty0 (16):\penalty0 1--6, 2016.

\bibitem[Larson and Chandler(2010)]{larson2010csiq}
Eric~Cooper Larson and Damon~Michael Chandler.
\newblock Most apparent distortion: full-reference image quality assessment and
  the role of strategy.
\newblock \emph{Journal of electronic imaging}, 19\penalty0 (1):\penalty0
  011006, 2010.

\bibitem[Li et~al.(2020)Li, Jiang, and Jiang]{li2020norm}
Dingquan Li, Tingting Jiang, and Ming Jiang.
\newblock Norm-in-norm loss with faster convergence and better performance for
  image quality assessment.
\newblock In \emph{ACM Multimedia}, 2020.

\bibitem[Li et~al.(2021)Li, Jiang, and Jiang]{li2021unified}
Dingquan Li, Tingting Jiang, and Ming Jiang.
\newblock Unified quality assessment of in-the-wild videos with mixed datasets
  training.
\newblock \emph{IJCV}, 129\penalty0 (4), 2021.

\bibitem[Li(2014)]{li2014learning}
Hang Li.
\newblock Learning to rank for information retrieval and natural language
  processing.
\newblock \emph{Synthesis lectures on human language technologies}, 7\penalty0
  (3):\penalty0 1--121, 2014.

\bibitem[Lin et~al.(2019)Lin, Hosu, and Saupe]{lin2019kadid}
Hanhe Lin, Vlad Hosu, and Dietmar Saupe.
\newblock Kadid-10k: A large-scale artificially distorted iqa database.
\newblock In \emph{QoMEX}, 2019.

\bibitem[Liu(2011)]{liu2011learning}
Tie-Yan Liu.
\newblock \emph{Learning to rank for information retrieval}.
\newblock Springer, 2011.

\bibitem[Liu et~al.(2018)Liu, Duanmu, and Wang]{liu2018end}
Wentao Liu, Zhengfang Duanmu, and Zhou Wang.
\newblock End-to-end blind quality assessment of compressed videos using deep
  neural networks.
\newblock In \emph{ACM Multimedia}, 2018.

\bibitem[Liu et~al.(2017)Liu, van De~Weijer, and Bagdanov]{liu2017rankiqa}
Xialei Liu, Joost van De~Weijer, and Andrew~D Bagdanov.
\newblock Rankiqa: Learning from rankings for no-reference image quality
  assessment.
\newblock In \emph{ICCV}, 2017.

\bibitem[Liu et~al.(2013)Liu, Wang, Cho, Finkelstein, and
  Rusinkiewicz]{liu2013deblurring}
Yiming Liu, Jue Wang, Sunghyun Cho, Adam Finkelstein, and Szymon Rusinkiewicz.
\newblock A no-reference metric for evaluating the quality of motion
  deblurring.
\newblock \emph{TOG}, 32\penalty0 (6):\penalty0 175--1, 2013.

\bibitem[Loshchilov and Hutter(2017)]{loshchilov2017sgdr}
Ilya Loshchilov and Frank Hutter.
\newblock Sgdr: Stochastic gradient descent with warm restarts.
\newblock In \emph{ICLR}, 2017.

\bibitem[Ma et~al.(2017)Ma, Yang, Yang, and Yang]{ma2017learning}
Chao Ma, Chih-Yuan Yang, Xiaokang Yang, and Ming-Hsuan Yang.
\newblock Learning a no-reference quality metric for single-image
  super-resolution.
\newblock \emph{CVIU}, 158:\penalty0 1--16, 2017.

\bibitem[Ma et~al.(2018)Ma, Duanmu, and Wang]{ma2018geometric}
Kede Ma, Zhengfang Duanmu, and Zhou Wang.
\newblock Geometric transformation invariant image quality assessment using
  convolutional neural networks.
\newblock In \emph{ICASSP}, 2018.

\bibitem[Manning and Raghavan(2008)]{manning2008ir}
Christopher~D Manning and Prabhakar Raghavan.
\newblock \emph{Introduction to Information Retrieval}.
\newblock Cambridge University Press, 2008.

\bibitem[Min et~al.(2019)Min, Zhai, Gu, Zhu, Zhou, Guo, Yang, Guan, and
  Zhang]{min2019dehazing}
Xiongkuo Min, Guangtao Zhai, Ke~Gu, Yucheng Zhu, Jiantao Zhou, Guodong Guo,
  Xiaokang Yang, Xinping Guan, and Wenjun Zhang.
\newblock Quality evaluation of image dehazing methods using synthetic hazy
  images.
\newblock \emph{TMM}, 21\penalty0 (9):\penalty0 2319--2333, 2019.

\bibitem[M{\"o}ller and Raake(2014)]{moller2014quality}
Sebastian M{\"o}ller and Alexander Raake.
\newblock \emph{Quality of experience: advanced concepts, applications and
  methods}.
\newblock Springer, 2014.

\bibitem[Murray et~al.(2012)Murray, Marchesotti, and Perronnin]{murray2012ava}
Naila Murray, Luca Marchesotti, and Florent Perronnin.
\newblock Ava: A large-scale database for aesthetic visual analysis.
\newblock In \emph{CVPR}, 2012.

\bibitem[Paszke et~al.(2019)Paszke, Gross, Massa, Lerer, Bradbury, Chanan,
  Killeen, Lin, Gimelshein, Antiga, et~al.]{paszke2019pytorch}
Adam Paszke, Sam Gross, Francisco Massa, Adam Lerer, James Bradbury, Gregory
  Chanan, Trevor Killeen, Zeming Lin, Natalia Gimelshein, Luca Antiga, et~al.
\newblock Pytorch: An imperative style, high-performance deep learning library.
\newblock \emph{Advances in neural information processing systems}, 32, 2019.

\bibitem[Pearson(1895)]{pearson1895vii}
Karl Pearson.
\newblock Vii. note on regression and inheritance in the case of two parents.
\newblock \emph{proceedings of the royal society of London}, 58\penalty0
  (347-352):\penalty0 240--242, 1895.

\bibitem[Ponomarenko et~al.(2013)Ponomarenko, Ieremeiev, Lukin, Egiazarian,
  Jin, Astola, Vozel, Chehdi, Carli, Battisti, et~al.]{ponomarenko2013color}
Nikolay Ponomarenko, Oleg Ieremeiev, Vladimir Lukin, Karen Egiazarian, Lina
  Jin, Jaakko Astola, Benoit Vozel, Kacem Chehdi, Marco Carli, Federica
  Battisti, et~al.
\newblock Color image database tid2013: Peculiarities and preliminary results.
\newblock In \emph{EUVIP}, 2013.

\bibitem[Ponomarenko et~al.(2015)Ponomarenko, Jin, Ieremeiev, Lukin,
  Egiazarian, Astola, Vozel, Chehdi, Carli, Battisti, and
  Jay]{ponomarenko2015tid}
Nikolay Ponomarenko, Lina Jin, Oleg Ieremeiev, Vladimir Lukin, Karen
  Egiazarian, Jaakko Astola, Benoit Vozel, Kacem Chehdi, Marco Carli, Federica
  Battisti, and Kuo C.-C. Jay.
\newblock Image database tid2013: Peculiarities, results and perspectives.
\newblock \emph{Signal processing: Image communication}, 30:\penalty0 57--77,
  2015.

\bibitem[Prashnani et~al.(2018)Prashnani, Cai, Mostofi, and
  Sen]{prashnani2018pieapp}
Ekta Prashnani, Hong Cai, Yasamin Mostofi, and Pradeep Sen.
\newblock Pieapp: Perceptual image-error assessment through pairwise
  preference.
\newblock In \emph{CVPR}, 2018.

\bibitem[Qin et~al.(2010)Qin, Liu, and Li]{qin2010general}
Tao Qin, Tie-Yan Liu, and Hang Li.
\newblock A general approximation framework for direct optimization of
  information retrieval measures.
\newblock \emph{Information retrieval}, 13\penalty0 (4):\penalty0 375--397,
  2010.

\bibitem[Russakovsky et~al.(2015)Russakovsky, Deng, Su, Krause, Satheesh, Ma,
  Huang, Karpathy, Khosla, Bernstein, et~al.]{russakovsky2015imagenet}
Olga Russakovsky, Jia Deng, Hao Su, Jonathan Krause, Sanjeev Satheesh, Sean Ma,
  Zhiheng Huang, Andrej Karpathy, Aditya Khosla, Michael Bernstein, et~al.
\newblock Imagenet large scale visual recognition challenge.
\newblock \emph{IJCV}, 115\penalty0 (3):\penalty0 211--252, 2015.

\bibitem[Sheikh and Bovik(2006)]{sheikh2006vif}
Hamid~R Sheikh and Alan~C Bovik.
\newblock Image information and visual quality.
\newblock \emph{TIP}, 15\penalty0 (2):\penalty0 430--444, 2006.

\bibitem[Sheikh et~al.(2003)Sheikh, Wang, Cormack, and Bovik]{sheikh2003live}
Hamid~R Sheikh, Zhou Wang, Lawrence Cormack, and Alan~C Bovik.
\newblock Live image quality assessment database.
\newblock http://live.ece.utexas.edu/research/quality, 2003.

\bibitem[Sheikh et~al.(2006)Sheikh, Sabir, and Bovik]{sheikh2006statistical}
Hamid~R Sheikh, Muhammad~F Sabir, and Alan~C Bovik.
\newblock A statistical evaluation of recent full reference image quality
  assessment algorithms.
\newblock \emph{TIP}, 15\penalty0 (11):\penalty0 3440--3451, 2006.

\bibitem[Siahaan et~al.(2016{\natexlab{a}})Siahaan, Hanjalic, and
  Redi]{siahaan2016augmenting}
Ernestasia Siahaan, Alan Hanjalic, and Judith~A Redi.
\newblock Augmenting blind image quality assessment using image semantics.
\newblock In \emph{ISM}, 2016{\natexlab{a}}.

\bibitem[Siahaan et~al.(2016{\natexlab{b}})Siahaan, Hanjalic, and
  Redi]{siahaan2016does}
Ernestasia Siahaan, Alan Hanjalic, and Judith~A Redi.
\newblock Does visual quality depend on semantics? a study on the relationship
  between impairment annoyance and image semantics at early attentive stages.
\newblock \emph{Electronic Imaging}, 2016\penalty0 (16):\penalty0 1--9,
  2016{\natexlab{b}}.

\bibitem[Siahaan et~al.(2018)Siahaan, Hanjalic, and Redi]{siahaan2018semantic}
Ernestasia Siahaan, Alan Hanjalic, and Judith~A Redi.
\newblock Semantic-aware blind image quality assessment.
\newblock \emph{Signal Processing: Image Communication}, 60:\penalty0 237--252,
  2018.

\bibitem[Simonyan and Zisserman(2015)]{simonyan2015vgg}
Karen Simonyan and Andrew Zisserman.
\newblock Very deep convolutional networks for large-scale image recognition.
\newblock In \emph{ICLR}, 2015.

\bibitem[Spearman(1904)]{spearman1904proof}
C~Spearman.
\newblock The proof and measurement of association between two things.
\newblock \emph{The American Journal of Psychology}, 15\penalty0 (1):\penalty0
  72--101, 1904.

\bibitem[Tian et~al.(2018)Tian, Zhang, Morin, and
  D{\'e}forges]{tian2018rendering}
Shishun Tian, Lu~Zhang, Luce Morin, and Olivier D{\'e}forges.
\newblock A benchmark of dibr synthesized view quality assessment metrics on a
  new database for immersive media applications.
\newblock \emph{TMM}, 21\penalty0 (5):\penalty0 1235--1247, 2018.

\bibitem[Wang et~al.(2018)Wang, Yu, Wu, Gu, Liu, Dong, Qiao, and
  Change~Loy]{wang2018esrgan}
Xintao Wang, Ke~Yu, Shixiang Wu, Jinjin Gu, Yihao Liu, Chao Dong, Yu~Qiao, and
  Chen Change~Loy.
\newblock Esrgan: Enhanced super-resolution generative adversarial networks.
\newblock In \emph{ECCVw}, 2018.

\bibitem[Wang and Bovik(2006)]{wang2006modern}
Zhou Wang and Alan~C Bovik.
\newblock \emph{Modern image quality assessment}, volume~2.
\newblock Morgan \& Claypool Publishers, 2006.

\bibitem[Wang et~al.(2003)Wang, Simoncelli, and Bovik]{wang2003msssim}
Zhou Wang, Eero~P Simoncelli, and Alan~C Bovik.
\newblock Multiscale structural similarity for image quality assessment.
\newblock In \emph{IEEE Asilomar Conference on Signals, Systems \& Computers},
  2003.

\bibitem[Wang et~al.(2004)Wang, Bovik, Sheikh, and Simoncelli]{wang2004image}
Zhou Wang, Alan~C Bovik, Hamid~R Sheikh, and Eero~P Simoncelli.
\newblock Image quality assessment: from error visibility to structural
  similarity.
\newblock \emph{TIP}, 13\penalty0 (4):\penalty0 600--612, 2004.

\bibitem[Watson and Null(1997)]{watson1997digital}
Andrew~B Watson and Cynthia~H Null.
\newblock Digital images and human vision.
\newblock In \emph{Electronic Imaging Science and Technology Conference}, 1997.

\bibitem[Xue et~al.(2013)Xue, Zhang, Mou, and Bovik]{xue2013gmsd}
Wufeng Xue, Lei Zhang, Xuanqin Mou, and Alan~C Bovik.
\newblock Gradient magnitude similarity deviation: A highly efficient
  perceptual image quality index.
\newblock \emph{TIP}, 23\penalty0 (2):\penalty0 684--695, 2013.

\bibitem[Zeyde et~al.(2010)Zeyde, Elad, and Protter]{zeyde2010single}
Roman Zeyde, Michael Elad, and Matan Protter.
\newblock On single image scale-up using sparse-representations.
\newblock In \emph{International conference on curves and surfaces}, pages
  711--730. Springer, 2010.

\bibitem[Zhang et~al.(2011)Zhang, Zhang, Mou, and Zhang]{zhang2011fsim}
Lin Zhang, Lei Zhang, Xuanqin Mou, and David Zhang.
\newblock Fsim: A feature similarity index for image quality assessment.
\newblock \emph{TIP}, 20\penalty0 (8):\penalty0 2378--2386, 2011.

\bibitem[Zhang et~al.(2014)Zhang, Shen, and Li]{zhang2014vsi}
Lin Zhang, Ying Shen, and Hongyu Li.
\newblock {VSI}: {A} visual saliency-induced index for perceptual image quality
  assessment.
\newblock \emph{TIP}, 23\penalty0 (10):\penalty0 4270--4281, 2014.

\bibitem[Zhang et~al.(2018)Zhang, Isola, Efros, Shechtman, and
  Wang]{zhang2018lpips}
Richard Zhang, Phillip Isola, Alexei~A Efros, Eli Shechtman, and Oliver Wang.
\newblock The unreasonable effectiveness of deep features as a perceptual
  metric.
\newblock In \emph{CVPR}, 2018.

\bibitem[Zhang et~al.(2021)Zhang, Ma, Zhai, and Yang]{zhang2021uncertainty}
Weixia Zhang, Kede Ma, Guangtao Zhai, and Xiaokang Yang.
\newblock Uncertainty-aware blind image quality assessment in the laboratory
  and wild.
\newblock \emph{TIP}, 30, 2021.

\bibitem[Zhu et~al.(2016)Zhu, Hanjalic, and Redi]{zhu2016qoe}
Yi~Zhu, Alan Hanjalic, and Judith~A Redi.
\newblock Qoe prediction for enriched assessment of individual video viewing
  experience.
\newblock In \emph{ACM Multimedia}, 2016.

\end{thebibliography}
